# Deep Learning of Subsurface Flow via Theory-guided Neural Network

Nanzhe Wang[a], Dongxiao Zhang[b,*], Haibin Chang[a], and Heng Li[a]

[a]BIC-ESAT, ERE, and SKLTCS, College of Engineering, Peking University, Beijing 100871, P. R. China
[b]School of Environmental Science and Engineering, Southern University of Science and Technology, Shenzhen 518055, P. R. China
[*]Corresponding author: E-mail address: zhangdx@sustech.edu.cn (Dongxiao Zhang)

## Abstract

Active researches are currently being performed to incorporate the wealth of scientific knowledge into data-driven approaches (e.g., neural networks) in order to improve the latter's effectiveness. In this study, the Theory-guided Neural Network (TgNN) is proposed for deep learning of subsurface flow. In the TgNN, as supervised learning, the neural network is trained with available observations or simulation data while being simultaneously guided by theory (e.g., governing equations, other physical constraints, engineering controls, and expert knowledge) of the underlying problem. The TgNN can achieve higher accuracy than the ordinary Artificial Neural Network (ANN) because the former provides physically feasible predictions and can be more readily generalized beyond the regimes covered with the training data. Furthermore, the TgNN model is proposed for subsurface flow with heterogeneous model parameters. Several numerical cases of two-dimensional transient saturated flow are introduced to test the performance of the TgNN. In the learning process, the loss function contains data mismatch, as well as PDE constraint, engineering control, and expert knowledge. After obtaining the parameters of the neural network by minimizing the loss function, a TgNN model is built that not only fits the data, but also adheres to physical/engineering constraints. Predicting the future response can be easily realized by the TgNN model. In addition, the TgNN model is tested in more complicated scenarios, such as prediction with changed boundary conditions, learning from noisy data or outliers, transfer learning, and engineering controls. Numerical results demonstrate that the TgNN model achieves much better predictability, reliability, and generalizability than ANN models due to the physical/engineering constraints in the former.

## 1. Introduction

In recent years, the Deep Neural Network (DNN) has gained increased attention and achieved great progress in artificial intelligence (AI), including image recognition (Krizhevsky et al., 2012; Simonyan & Zisserman, 2014), natural language processing (Collobert & Weston, 2008; Goldberg, 2016), automatic driving (Dong et al., 2018; Manikandan & Ganesan, 2019; Wang et al., 2018), speech recognition (Li et al., 2018; Novoa et al., 2018), etc. The tremendous advances of DNN benefited from the development of computer hardware, such as graphics processing unit (GPU), which provides powerful computing resources and speeds up calculations.

In addition to the field of AI, DNN has also been utilized in diverse scientific disciplines, including the fields of biomedicine (Liang et al., 2018; Shashikumar et al., 2018), economics (Singh & Srivastava, 2017; Yong et al., 2017), chemistry (Fooshee et al., 2018; Sun et al., 2019), and physics (Bhimji et al., 2018; Sadowski & Baldi, 2018). Despite the numerous successes obtained with DNN, limitations remain concerning the application of DNN in numerous scientific problems due to the following reasons: First, a large amount of data is usually requisite to guarantee model accuracy. The constitutes a major challenge because, in many scientific and engineering practices, data collection is both time-consuming and expensive. Moreover, without sufficient data, DNN may exhibit low reliability and a poor ability to generalize beyond the regimes covered with the training data. Second, the DNN model is only trained by the available dataset, and no scientific principles or laws are involved during the model training, which may lead to physically unreasonable predictions for some specific scientific problems. Third, the quality of collected data may not be ensured in practical measurement, and thus noise or outliers may exist in the dataset. Indeed, DNN may experience extreme interference due to noise or outliers, thus producing completely incorrect results.

To overcome these limitations, incorporating scientific knowledge or practical experience into deep learning models is an emerging paradigm for many scientific problems. For example, Karpatne et al. (2017) proposed a Theory-Guided Data Science (TGDS) approach, which integrates scientific knowledge and data science. In their work, five ways are presented to achieve integration, which are theory-guided design of data science models, theory-guided learning of data science models, theory-guided refinement of data science outputs, learning hybrid models of theory and data science, and augmenting theory-based models utilizing data science. Karpatne et al. (2017) also proposed a physics-guided neural network (PGNN) model, which adds physics-based loss into the learning objective function of neural network to obtain scientifically consistent results. Moreover, the proposed PGNN is applied to a lake temperature modeling problem. Raissi et al. (2019) proposed Physics-Informed Neural Networks (PINN), in which a constraint term from physical laws described by general nonlinear partial differential equations is incorporated into the neural network training. The PINN can be used to realize data-driven solutions and inverse modeling of partial differential equations.

More generally, in this work, we propose the Theory-guided Neural Network (TgNN) framework, which can incorporate the theory (e.g., governing equations, other physical constraints, engineering controls, and expert knowledge) of the underlying problem into neural network training. Scientific laws and engineering theories, serving as prior knowledge, are transformed into regularization terms and added into the loss function, which can assist to guide the training of the DNN. Consequently, the TgNN can achieve higher accuracy than the (deep) artificial neural network (ANN) because the former provides physically feasible predictions and can be more readily generalized beyond the regimes covered with the training data. In this work, the proposed TgNN framework is employed to deal with subsurface flow with heterogeneous model parameters. Several illustrative two-dimensional subsurface problems with different scenarios, including changing boundary conditions, training from noisy data or outliers, and transfer learning, are designed to test the performance of the proposed TgNN. By comparison with the conventional ANN, TgNN achieves superior performance.

The remainder of this paper proceeds as follows. In section 2, we briefly introduce the architecture and mechanism of DNN, and present the framework of TgNN. In section 3, several illustrative subsurface flow cases with different scenarios are designed to test performance of the TgNN. Finally, discussions and conclusions are provided in section 4.

## 2. Methodology

In this section, we first briefly introduce the architecture of the deep neural network. Then, we elaborate on the concept of Theory-guided Neural Network (TgNN). Finally, we investigate how to incorporate the governing equation constraint into TgNN when the model parameter is heterogeneous.

### 2.1 Deep Neural Network

The Deep Neural Network (DNN) is a powerful function approximation tool, which can learn the relationship between input and output variables. There is an input layer, an output layer, and hidden layers in the neural network architecture, each of which consists of several neurons, as shown in **Figure 1**. A DNN usually has more than one hidden layer. For simplicity, let us assume that there are L hidden layers, the input is a vector $\mathbf{X}$, and the output is a vector $\mathbf{Y}$. The forward formulation of DNN can then be represented as follows:

$$\begin{aligned}
\mathbf{z}^{1} &= \sigma_1(\mathbf{W}^{1}\mathbf{X}+\mathbf{b}^{1}) \\
\mathbf{z}^{2} &= \sigma_2(\mathbf{W}^{2}\mathbf{z}^{1}+\mathbf{b}^{2}) \\
&\vdots \\
\mathbf{z}^{L} &= \sigma_L(\mathbf{W}^{L}\mathbf{z}^{L-1}+\mathbf{b}^{L}) \\
\mathbf{Y} &= \mathbf{W}^{L+1}\mathbf{z}^{L}+\mathbf{b}^{L}
\end{aligned} \tag{1}$$

where $\mathbf{W^i}$ and $\mathbf{b^i}$ are weights and bias of the $i^{\text{th}}$ layer, respectively, which are known as network parameters; $\theta=\{\mathbf{W^i},\mathbf{b^i}\}_{i=1}^{L+1}$ (here, superscript L+1 denotes the output layer); and $\sigma^i$ is the activation function of the $i^{\text{th}}$ layer, such as Sigmoid, hyperbolic tangent (Tanh), and Rectified Linear Unit (ReLU) (Goodfellow et al., 2016). The forward formulation in **Eq. (1)** can be simply expressed as $\mathbf{Y}=N(\mathbf{X};\theta)$. Then, the loss function, which is usually the mean square error between the output and the ground truth data, can be represented as:

$$L(\theta)=MSE_{DATA}=\frac{1}{N}\sum_{i=1}^{N}\left|N(x_i;\theta)-y_i\right|^2 \tag{2}$$

where $N$ denotes the total number of labeled data. During the training process of DNN, the network parameters can be tuned by minimizing the loss function via an optimization algorithm, such as Stochastic gradient descent (Bottou, 2010). The trained DNN can then be used to obtain prediction for the new inputs.

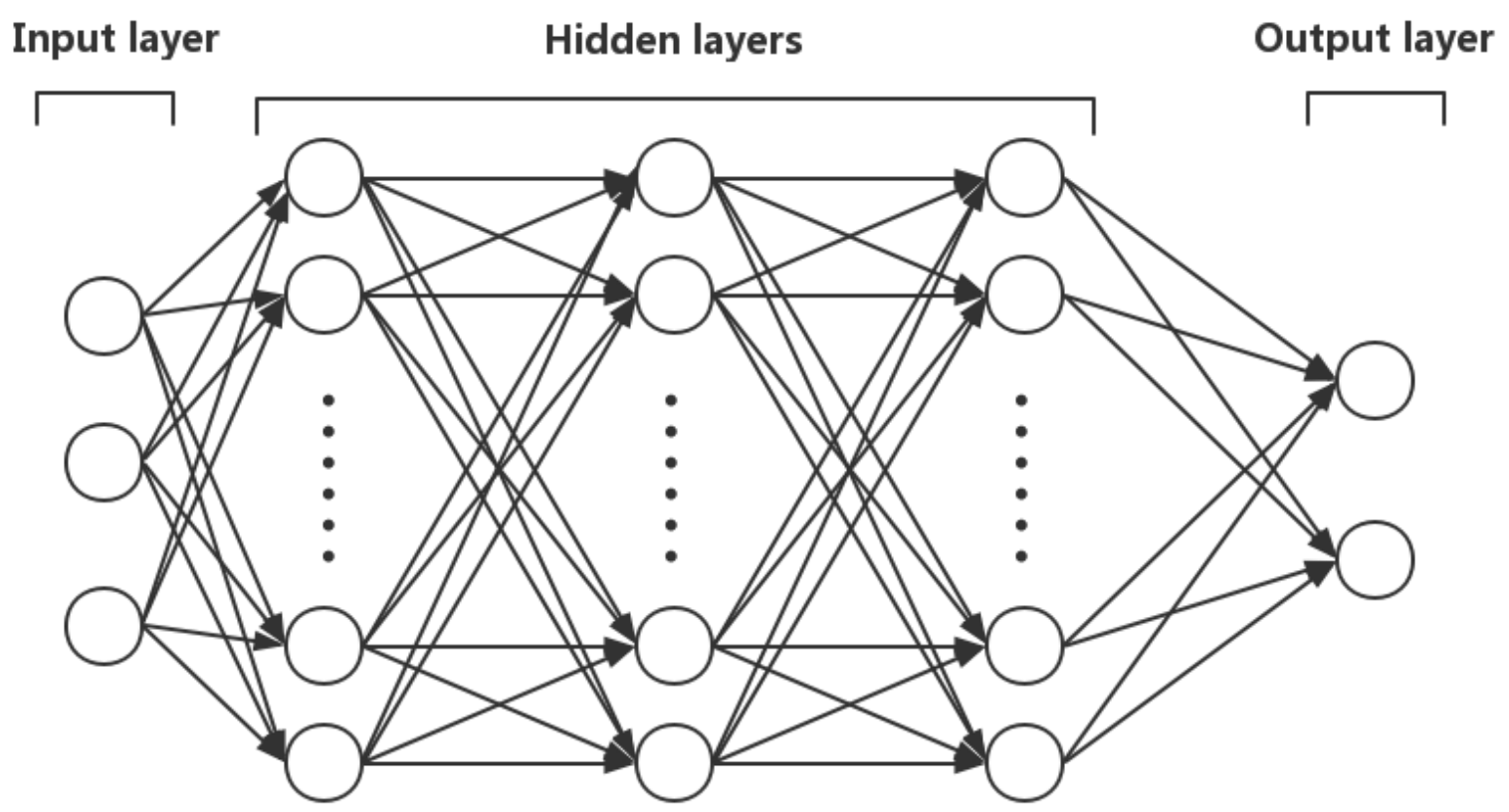

**Figure 1.** Architecture of Neural Network

## 2.2 Theory-guided Neural Network

For DNN, a large amount of data may be required for approximating complex functions to achieve desirable accuracy. However, in many scientific or engineering practices, data acquisition may be difficult and time-consuming, and thus the available data may be scarce. On the other hand, DNN may produce physically unreasonable predictions for a specific scientific problem without incorporating scientific laws and practical theories. Since the response of a physical problem should obey the theory (e.g., governing equations, other physical constraints, engineering controls, and expert knowledge) of the underlying problem, these theories, constituting prior knowledge, may be utilized to guide the DNN training to lessen the data dependence of the DNN. In the following, we will elaborate on how to construct the Theory-guided Neural Network (TgNN).

We consider that a subsurface flow in saturated homogeneous porous medium satisfies the following governing equation:

$$S_s \frac{\partial h}{\partial t} = K \frac{\partial^2 h}{\partial x^2} + K \frac{\partial^2 h}{\partial y^2} \tag{3}$$

where $S_s$ denotes specific storage; $K$ denotes hydraulic conductivity; and $h$ denotes the hydraulic head. The boundary conditions and initial conditions can be expressed as follows:

$$h(x_{BC}, y_{BC}) = h_{BC}, \quad h(t_{IC}) = h_{IC} \tag{4}$$

Denote the hydraulic head as $h(t,x,y)$, and in the TgNN model, it is approximated by a neural network as $N_h(t,x,y;\theta)$. Considering that $N_h(t,x,y;\theta)$ should also follow the governing equation, we define a function, which represents the residual of the governing equation, as:

$$f := S_s \frac{\partial N_h(t,x,y;\theta)}{\partial t} - K \frac{\partial^2 N_h(t,x,y;\theta)}{\partial x^2} - K \frac{\partial^2 N_h(t,x,y;\theta)}{\partial y^2} \tag{5}$$

where the partial derivatives can be easily computed by applying the chain rule for the network through automatic differentiation, which can be easily implemented in the deep learning framework, such as Pytorch (Paszke et al., 2017), tensorflow (Abadi et al., 2016), etc. In order to enforce the governing equation constraint during the training process, $f$ needs to approach to zero, and thus the mean squared error of $f$ should be included in the loss function as:

$$MSE_{PDE} = \frac{1}{N_f} \sum_{i=1}^{N_f} \left| f(t_f^i, x_f^i, y_f^i) \right|^2 \tag{6}$$

where $\left\{ t_f^i, x_f^i, y_f^i \right\}_{i=1}^{N_f}$ denotes the collocation points of the residual function, which can be randomly chosen because no labels are needed for these points. Moreover, the boundary conditions and initial conditions of the dynamical system can also be expressed in residual form, as shown below:

$$f_{BC} := N_h(t, x_{BC}, y_{BC}; \theta) - h_{BC} \tag{7}$$

$$f_{IC} := N_h(t_{IC}, x, y; \theta) - h_{IC} \tag{8}$$

The mean squared error of boundary and initial conditions can then be written as:

$$MSE_{BC} = \frac{1}{N_{BC}} \sum_{i=1}^{N_{BC}} \left| f_{BC}(t^i, x_{BC}^i, y_{BC}^i) \right|^2 \tag{9}$$

$$MSE_{IC} = \frac{1}{N_{IC}} \sum_{i=1}^{N_{IC}} \left| f_{IC}(t_{IC}^i, x^i, y^i) \right|^2 \tag{10}$$

Additionally, some engineering controls and expert knowledge in practice may also assist to guide the construction of the model, for example, the hydraulic head value should fall in a certain range due to some specific boundary and initial conditions, or the pumping rate of a well may be expected to be less than some value, all of which may be incorporated into the training process. Without loss of generality, the engineering controls and expert knowledge in the system can be expressed as follows:

$$EC(t, x, y) \le 0 \tag{11}$$

$$EK(t, x, y) \le 0 \tag{12}$$

When those controls are violated, there should be a penalty term reflected in the loss function, and the corresponding penalty functions for engineering controls and expert knowledge can be introduced as:

$$f_{EC} = \text{ReLU}(EC(t^i, x^i, y^i)) \tag{13}$$

$$f_{EK} = \text{ReLU}(EK(t^i, x^i, y^i)) \tag{14}$$

where ReLU(.) is the rectified linear unit function. The ReLU function returns zero when the inputs are negative, and returns the original value of inputs when they are positive. Consequently, it has the same mechanism as max(0, x). Then, the loss terms of engineering controls and expert knowledge are shown as follows:

$$MSE_{EC} = \frac{1}{N_{EC}} \sum_{i=1}^{N_{EC}} \left| \text{ReLU}(EC(t^i, x^i, y^i)) \right|^2 \tag{15}$$

$$MSE_{EK} = \frac{1}{N_{EK}} \sum_{i=1}^{N_{EK}} \left| \text{ReLU}(EK(t^i, x^i, y^i)) \right|^2 \tag{16}$$

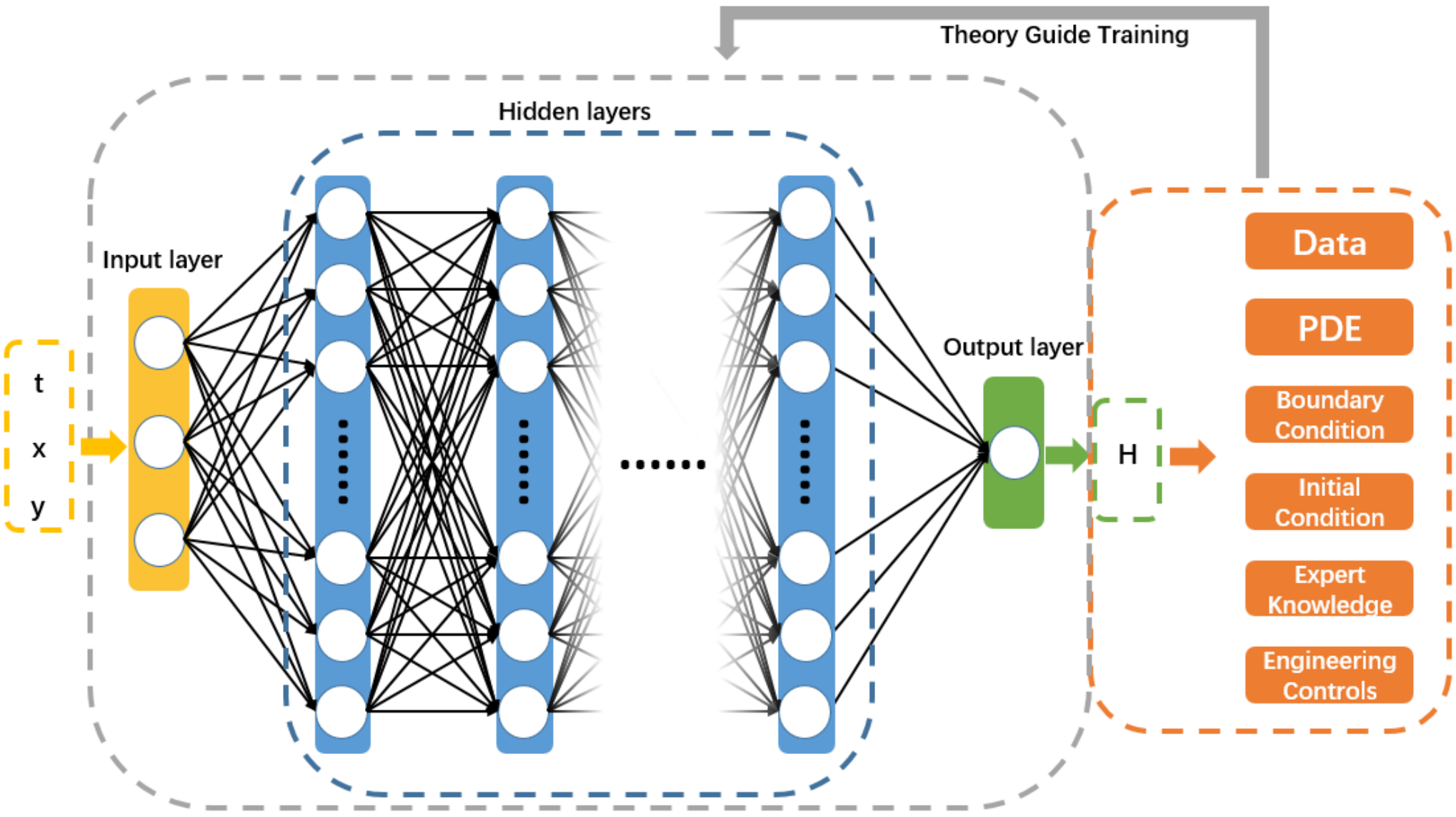

**Figure 2.** Structure of the TgNN model.

Thus, the loss function of the TgNN model, which incorporates scientific knowledge and engineering controls, can be written as:

$$\begin{aligned} L(\theta) = {} & \lambda_{DATA} MSE_{DATA} + \lambda_{PDE} MSE_{PDE} + \lambda_{BC} MSE_{BC} \\ & + \lambda_{IC} MSE_{IC} + \lambda_{EC} MSE_{EC} + \lambda_{EK} MSE_{EK} \end{aligned} \tag{17}$$

where $\lambda_{DATA}$, $\lambda_{PDE}$, $\lambda_{BC}$, $\lambda_{IC}$, $\lambda_{EC}$, and $\lambda_{EK}$ are the hyper-parameters, which control the weight of each term in the loss function.

The TgNN model can then be trained by minimizing the loss function via some optimization algorithm, such as Stochastic Gradient Descent (SGD), Adagrad, Adaptive Moment Estimation (Adam) (Kingma & Ba, 2015), etc. These optimization algorithms have been commonly and successfully employed in the training process of DNN. The structure of the TgNN model is illustrated in **Figure 2.**

### 2.3 Governing equation with heterogeneous model parameter

For subsurface flow problems, the parameter fields are usually heterogeneous. With heterogeneous hydraulic conductivity, the governing equation of flow in the saturated porous medium can be rewritten as follows:

$$S_s \frac{\partial h}{\partial t} = \frac{\partial}{\partial x}\left( K(x,y)\frac{\partial h}{\partial x} \right) + \frac{\partial}{\partial y}\left( K(x,y)\frac{\partial h}{\partial y} \right) \tag{18}$$

Then, the residual of the governing equation can be expressed as:

$$\begin{aligned} f := S_S \frac{\partial N_h\left(t,x,y;\theta\right)}{\partial t} - \frac{\partial}{\partial x}\left( K(x,y)\bullet\frac{\partial N_h\left(t,x,y;\theta\right)}{\partial x} \right) \\ - \frac{\partial}{\partial y}\left( K(x,y)\bullet\frac{\partial N_h\left(t,x,y;\theta\right)}{\partial y} \right) \end{aligned} \tag{19}$$

where the partial derivatives of $N_h\left(t,x,y;\theta\right)$ can be easily computed by applying the chain rule for the network through automatic differentiation, while the partial derivatives of $K(x,y)$ need to be calculated, in general, through numerical difference.

Considering that a heterogeneous parameter field can be treated as a realization of a random field following a specific distribution with corresponding covariance, Karhunen–Loeve expansion (KLE) can be introduced to parameterize the heterogeneous model parameter if its covariance is known.

For a random field $Z(\mathbf{x},\tau) = \ln K(\mathbf{x},\tau)$, where $\mathbf{x} \in D$ (physical domain) and $\tau \in \Theta$ (a probability space), it can be expressed as $Z(\mathbf{x},\tau) = \overline{Z}(\mathbf{x}) + Z'(\mathbf{x},\tau)$, where $\overline{Z}(\mathbf{x})$ is the mean of the random field, and $Z'(\mathbf{x},\theta)$ is the fluctuation. The spatial structure of the random field can be described by the covariance $C_Z(\mathbf{x},\mathbf{x}') = \left\langle Z'(\mathbf{x},\tau)Z'(\mathbf{x}',\tau) \right\rangle$. Since the covariance is bounded, symmetric, and positive-definite, it can be decomposed as (Ghanem & Spanos, 2003):

$$C_Z(\mathbf{x},\mathbf{x}') = \sum_{i=1}^{\infty} \lambda_i f_i(\mathbf{x}) f_i(\mathbf{x}') \tag{20}$$

where $\lambda_i$ and $f_i(\mathbf{x})$ are the eigenvalue and eigenfunction, respectively, which can be obtained by solving the second-type Fredholm equation:

$$\int_D C_Z(\mathbf{x},\mathbf{x}') f(\mathbf{x}) = \lambda f(\mathbf{x}') \tag{21}$$

Therefore, the random field can be expressed as follows:

$$Z(\mathbf{x},\tau) = \overline{Z}(\mathbf{x}) + \sum_{i=1}^{\infty} \sqrt{\lambda_i} f_i(\mathbf{x}) \xi_i(\tau) \tag{22}$$

where $\xi_i(\tau)$ are orthogonal Gaussian random variables with zero mean and unit variance.

Although there are infinite terms in **Eq. (22)**, we can truncate the expansion into a finite number of terms. The number of retained terms in the KL expansion should be determined by the decay rate of $\lambda_i$. Moreover, the number of retained terms determines the random dimensionality, for example, *n*. By using the KLE, the random field $Z(\mathbf{x},\tau)=\ln K(\mathbf{x},\tau)$ can be parameterized by a group of independent random variables as:

$$\xi=\left\{\xi_1(\tau),\xi_2(\tau),\cdots,\xi_n(\tau)\right\} \tag{23}$$

The random field can then be represented by:

$$Z(x,y)\approx\overline{Z(x,y)}+\sum_{i=1}^{n}\sqrt{\lambda_i}\,f_i(x,y)\xi_i(\tau) \tag{24}$$

Furthermore, the residual of the governing equation can be rewritten as follows:

$$\begin{aligned} f := S_s\frac{\partial N_h(t,x,y;\theta)}{\partial t}&-\frac{\partial}{\partial x}\left(e^{\overline{Z(x,y)}+\sum_{i=1}^{n}\sqrt{\lambda_i}f_i(x,y)\xi_i(\tau)}\bullet\frac{\partial N_h(t,x,y;\theta)}{\partial x}\right)\\ &-\frac{\partial}{\partial y}\left(e^{\overline{Z(x,y)}+\sum_{i=1}^{n}\sqrt{\lambda_i}f_i(x,y)\xi_i(\tau)}\bullet\frac{\partial N_h(t,x,y;\theta)}{\partial y}\right)\end{aligned} \tag{25}$$

In general, the eigenvalues and eigenfunctions should be solved numerically using **Eq. (21)** (Chang & Zhang, 2015). However, for the separable exponential covariance model, the eigenvalues and eigenfunctions can be solved analytically or semi-analytically, the details of which are given in **Appendix A**. In this work, for the two-dimensional parameter field, the covariance function is assumed to be a separable exponential model. Under this condition, the partial derivatives of $K(x, y)$ can be obtained analytically.

## 3. Case Studies

In this section, the performance of the proposed TgNN model is tested by several cases of subsurface flow. The accuracy and robustness of TgNN model is also compared with the ANN model in different situations.

A two-dimensional transient saturated flow in porous medium is considered, which satisfies the following governing equation:

$$S_s\frac{\partial h}{\partial t}=\frac{\partial}{\partial x}\left(K(x,y)\frac{\partial h}{\partial x}\right)+\frac{\partial}{\partial y}\left(K(x,y)\frac{\partial h}{\partial y}\right) \tag{26}$$

where *Ss* denotes the specific storage; *h* denotes the hydraulic head; and $K(x,y)$ denotes the hydraulic conductivity. The domain is a square, which is evenly divided into 51×51 grid blocks, and the length in both directions is $1020\,[L]$, where $[L]$ denotes any consistent length unit. Prescribed heads are assigned for the left and right boundaries, taking values of $H_{x=0}=1[L]$ and $H_{x=1020}=0[L]$, respectively. The two lateral boundaries are assigned as no-flow boundaries. The specific storage is assumed as a constant, taking a value of $S_s=0.0001[L^{-1}]$. The total simulation time is $10\,[T]$, where $[T]$ denotes any consistent time unit, with each time step being $0.2\,[T]$, resulting in 50 time steps. The initial conditions are $H_{t=0,x=0}=1[L]$ and $H_{t=0,x\neq 0}=0[L]$. The mean and variance of the log hydraulic conductivity are given as $\langle \ln K \rangle = 0$ and $\sigma_K^2=1.0$, respectively. In addition, the correlation length of the field is $\eta=408\,[L]$. The hydraulic conductivity field is parameterized through KLE, and 20 terms are retained in the expansion. Thus, this field is represented by 20 random variables $\xi=\{\xi_1(\tau),\xi_2(\tau),\cdots,\xi_{20}(\tau)\}$ in the considered cases. The MODFLOW software is adopted to perform the simulations to obtain the required dataset.

### 3.1 Predicting the future response

In this case, it is assumed that the hydraulic head distribution at the first 18 time steps is monitored, and 1000 data points are extracted at each time step as training data. We aim to predict the hydraulic head distribution at the following 32 time steps.

Considering that the hydraulic head is a function of time and space, i.e., $H(t,x,y)$, the inputs of the TgNN model are time ($t$) and position ($x,y$), and the output is hydraulic head ($H$). A fully-connected neural network with 7 hidden layers is used, which has 50 neurons in each hidden layer. In the loss function, the data term, PDE term, boundary condition term, and initial condition term are arranged as that in **Eq. (17)**. Regarding the expert knowledge term, we incorporate that the hydraulic head should take a value from 0 to 1 in this case, which can be expressed as follows:

$$N_h\left(t,x,y;\theta\right)-1\leq 0 \tag{27}$$

$$0-N_h\left(t,x,y;\theta\right)\leq 0 \tag{28}$$

Therefore, the expert knowledge term in the loss function is:

$$MSE_{EK} = \frac{1}{N_{EK}} \left( \sum_{i=1}^{N_{EK}} \left| \mathrm{ReLU}(N_h(t,x,y;\theta) - 1) \right|^2 + \sum_{i=1}^{N_{EK}} \left| \mathrm{ReLU}(0 - N_h(t,x,y;\theta)) \right|^2 \right) \tag{29}$$

In this way, in addition to the training data, the scientific knowledge is incorporated into the TgNN, which can assist to achieve scientific consistency of the model. The model structure is presented in **Figure 2**. The Adam algorithm is used to perform the optimization in the training process, which can compute adaptive learning rates for each parameter (Kingma & Ba, 2015).

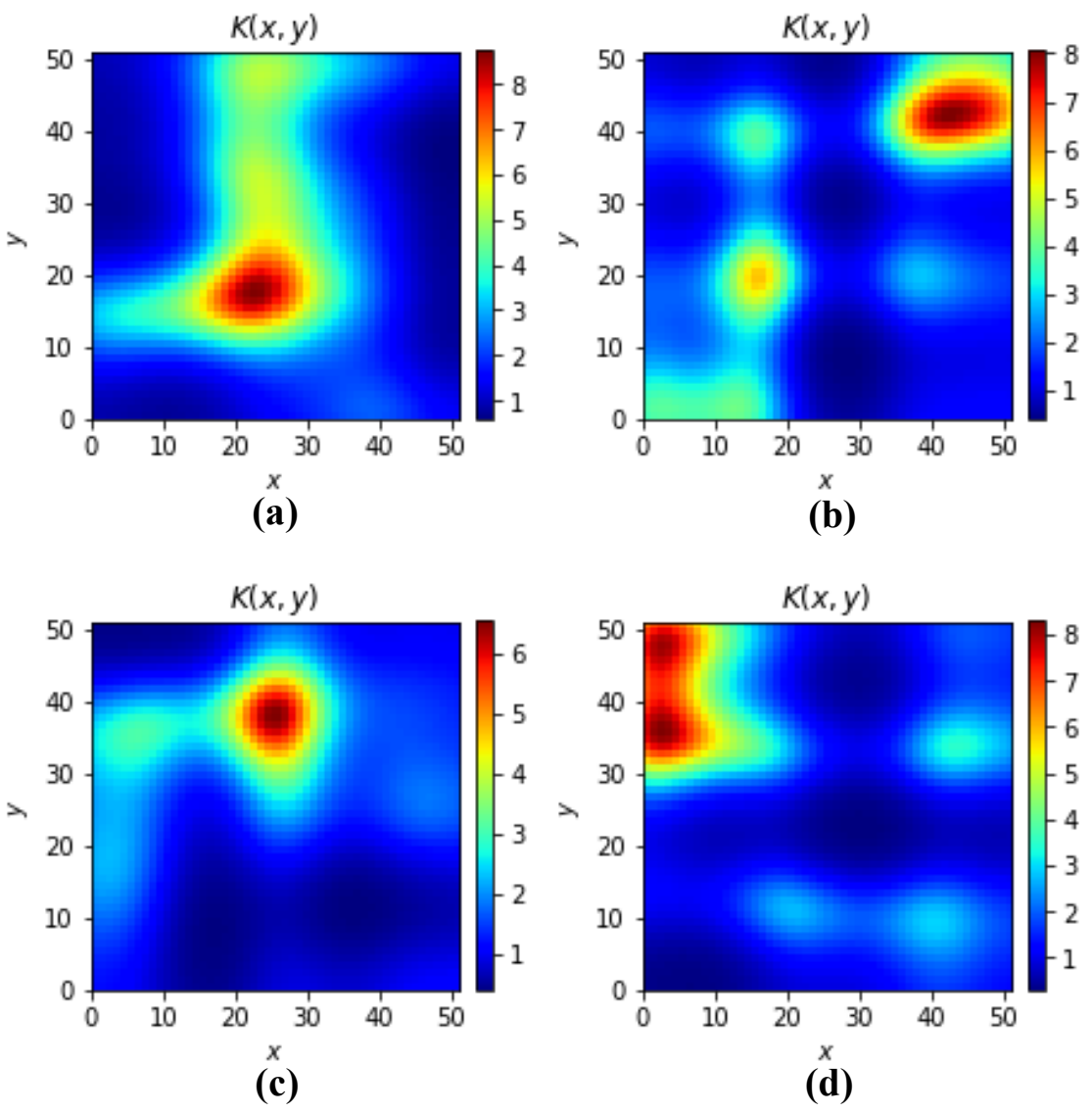

**Figure 3.** Four hydraulic conductivity fields (a, b, c, and d).

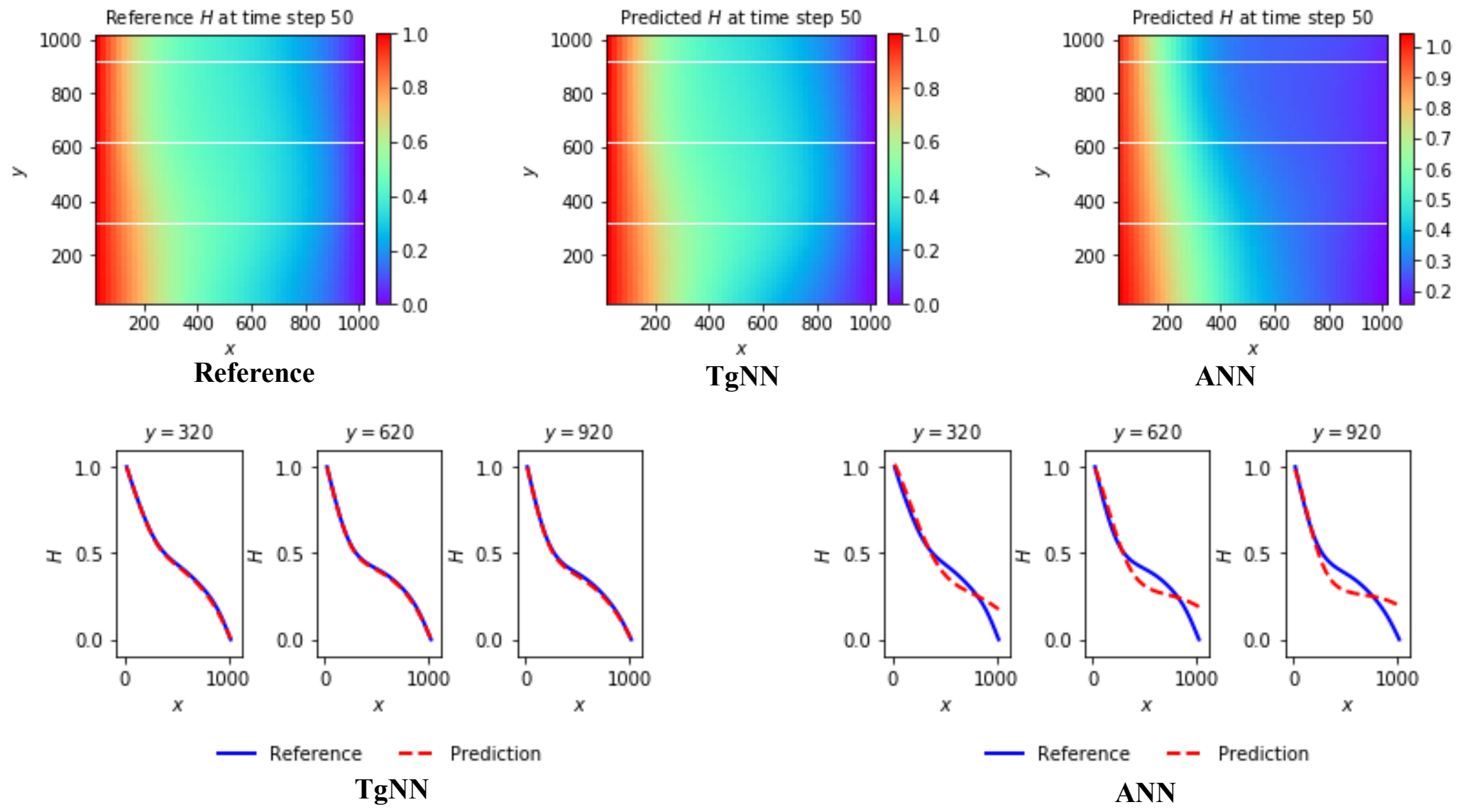

**Figure 4.** Prediction results of TgNN and ANN for hydraulic conductivity field (a).

The performance of the proposed approach is assessed using four different hydraulic conductivity fields, as shown in **Figure 3**. The predictions from TgNN and ANN at time step 50 (the last step) for hydraulic conductivity field (a) are presented in **Figure 4**, and the predictions for hydraulic conductivity field (b), (c), and (d) are shown in **Appendix B.1**. It can be seen that the predictions of TgNN match the reference values well and are superior to the predictions of ANN.

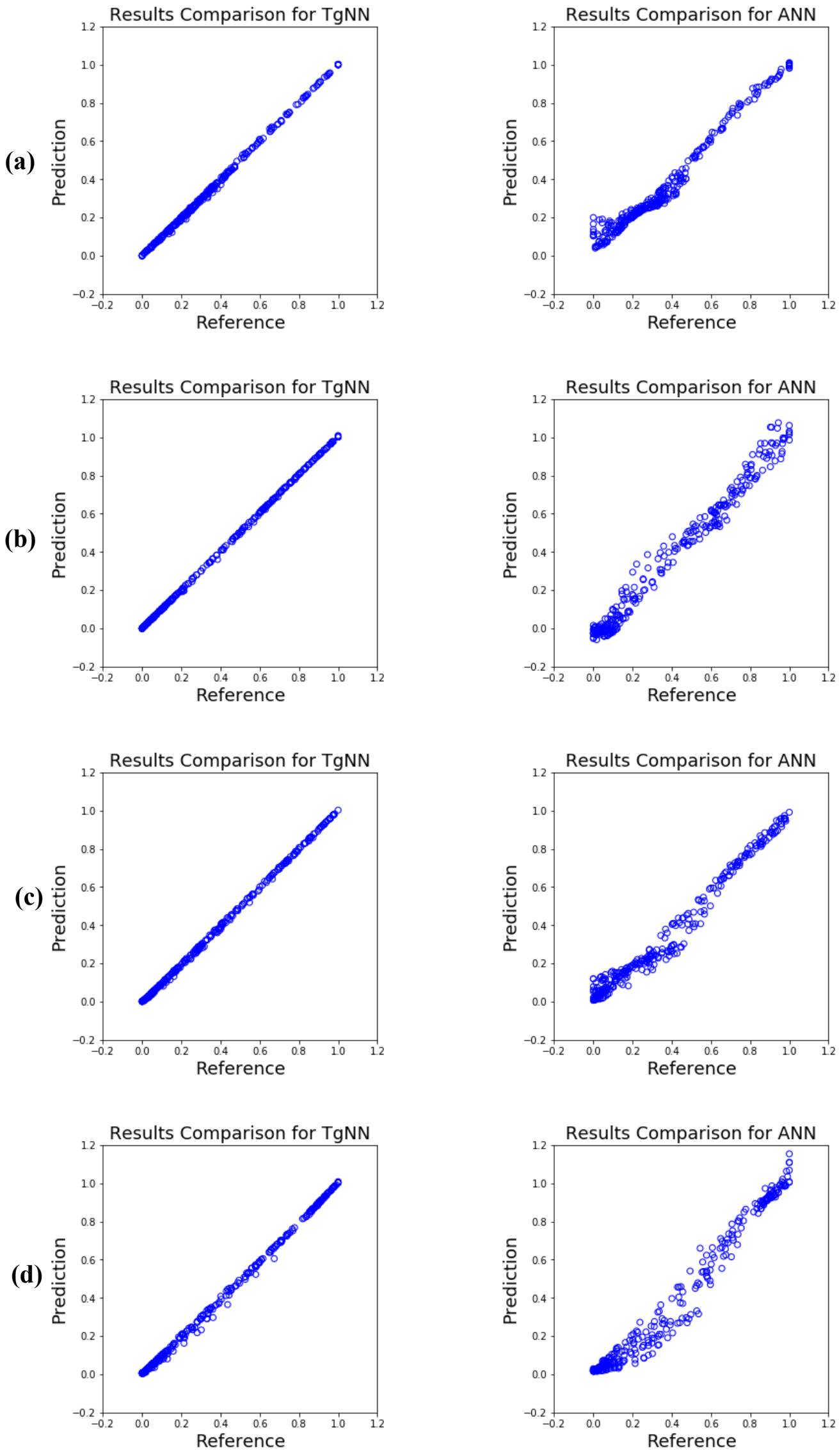

**Figure 5.** Correlation between the reference and predicted hydraulic head for four hydraulic conductivity fields.

Furthermore, in order to quantitatively evaluate the results, the following two metrics are introduced. One is the relative $L_2$ error:

$$L_2(u_{pred}, u_{true}) = \frac{\left\| u_{pred} - u_{true} \right\|_2}{\left\| u_{true} \right\|_2} \tag{30}$$

where $\| \cdot \|_2$ denotes the standard Euclidean norm; and $u_{true}$ and $u_{pred}$ are the reference value solved by MODFLOW and the predictions of the discussed two models, respectively. The other metric is the coefficient of determination, also known as $R^2$ score, which is defined as follows:

$$R^2 = 1 - \frac{\sum_{n=1}^{N_{cell}} (u_{pred,n} - u_{true,n})^2}{\sum_{n=1}^{N_{cell}} (u_{true,n} - \bar{u}_{true})^2} \tag{31}$$

where $N_{cell}$ denotes all of the blocks that need to be predicted; $u_{true,n}$ and $u_{pred,n}$ are the reference value solved by MODFLOW and the predictions of the discussed two models at the $n$th block, respectively; and $\bar{u}_{true}$ denotes the mean of $u_{true,n}$. The corresponding relative $L_2$ error and $R^2$ score for the predictions of TgNN and ANN are shown in **Table 1**.

**Table 1.** The relative $L_2$ error and $R^2$ score of predictions.

| | **Relative $L_2$ error** | | **$R^2$ score** | |
|---|---|---|---|---|
| | **TgNN** | **ANN** | **TgNN** | **ANN** |
| **K1** | 1.037615e-02 | 1.182732e-01 | 9.997364e-01 | 9.657531e-01 |
| **K2** | 1.721577e-02 | 9.557852e-02 | 9.991099e-01 | 9.725646e-01 |
| **K3** | 1.883147e-02 | 1.104504e-01 | 9.991939e-01 | 9.722684e-01 |
| **K4** | 2.531426e-02 | 1.213925e-01 | 9.984961e-01 | 9.654165e-01 |

Here, we tested 500 different realizations of the hydraulic conductivity fields generated by KLE, and all of the relative $L_2$ error and $R^2$ scores of corresponding predictions are statistically analyzed. **Figure 6** presents the histograms of the relative $L_2$ error and $R^2$ score for the 500 realizations, and **Table 2** shows their mean and variance. It can be seen that the relative $L_2$ error of the TgNN model is far less than that of the ANN model, and the $R^2$ score of the TgNN model is closer to 1 than the ANN model. Both of these findings indicate that the predicted hydraulic heads from the TgNN match the references better than those from the ANN.

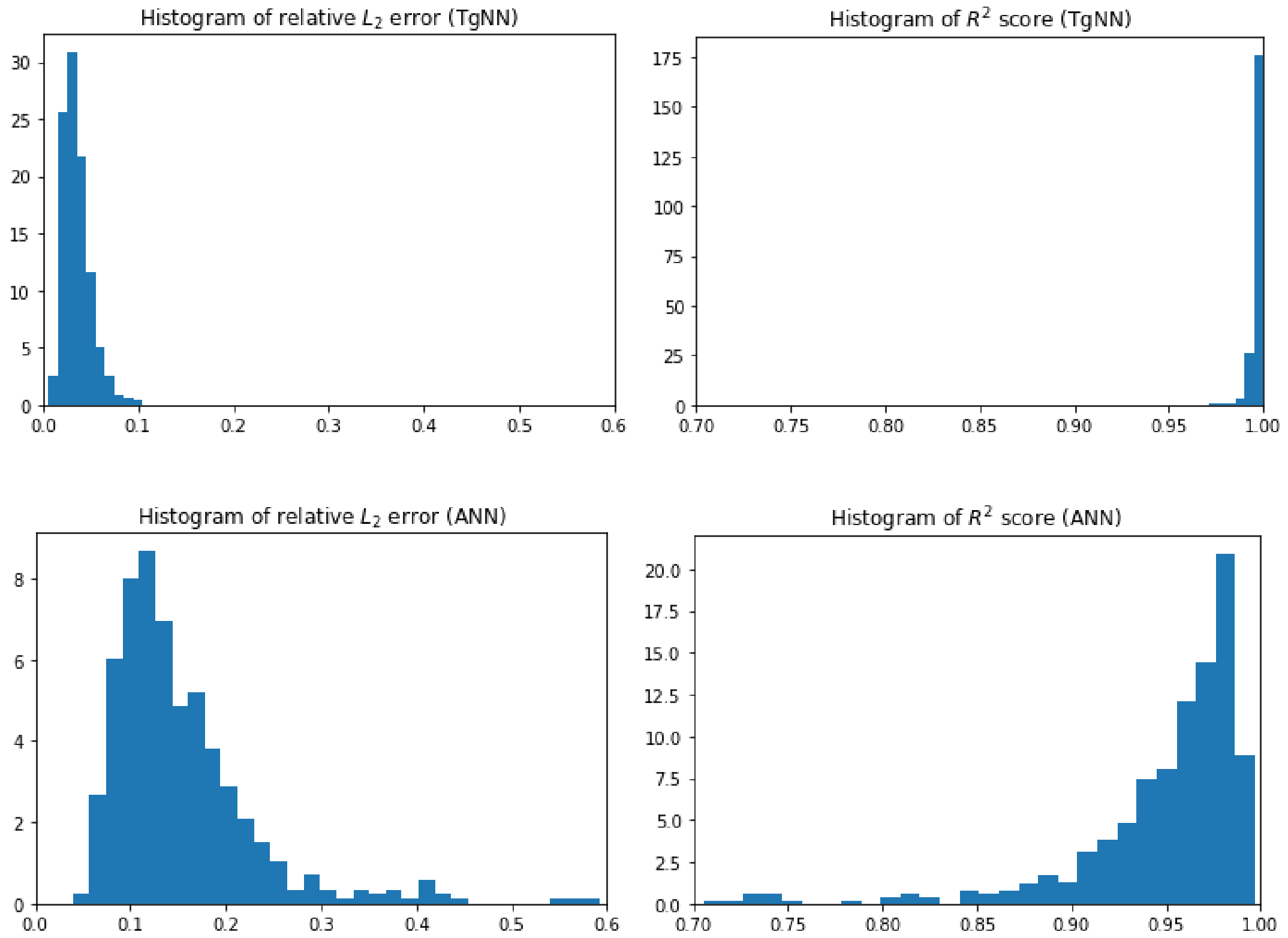

**Figure 6.** Histograms of relative $L_2$ error and $R^2$ score for 500 realizations.

**Table 2.** The mean and variance of relative $L_2$ error and $R^2$ score for 500 realizations.

| | **Relative $L_2$ error** | | **$R^2$ score** | |
|---|---|---|---|---|
| | **TgNN** | **ANN** | **TgNN** | **ANN** |
| **Mean** | 3.509975e-02 | 1.543908e-01 | 9.970567e-01 | 9.366969e-01 |
| **Variance** | 2.141023e-04 | 6.443073e-03 | 8.733067e-06 | 9.098723e-03 |

### 3.2 Predicting the future response with changed boundary conditions

In this subsection, the TgNN model is further tested by solving a more difficult situation, in which the boundary condition changes. In this case, the hydraulic head at the end $x = 1020[L]$ rises to $2[L]$ from $0[L]$ at time step 20, i.e., $t = 4[T]$. The hydraulic heads at the first 20 time steps are monitored, and 1000 data points are selected at each time step as training data. Since the data after the boundary condition changes are not available, the difficulty for predicting the hydraulic head afterwards will be increased. On the other hand, TgNN can straightforwardly handle this situation by incorporating the information of the new boundary condition into the loss function as follows:

$$MSE_{NEW-BC} = \frac{1}{N_{NEW-BC}} \sum_{i=1}^{N_{NEW-BC}} \left| f_{NEW-BC}(t^i_{NEW-BC}, x^i_{NEW-BC}, y^i_{NEW-BC}) \right|^2 \quad (32)$$

The second hydraulic conductivity field, shown in **Figure 3,** is chosen for performing this case study. **Figure 7** presents the correlation between the prediction and reference data, and **Appendix B.2** shows the predictions of the hydraulic head by TgNN and ANN at time step 30, 40, and 50, respectively. **Table 3** presents the relative $L_2$ error and $R^2$ score of the prediction. Compared with TgNN, ANN provides unsatisfactory predictions even though the new boundary condition is added into the training data.

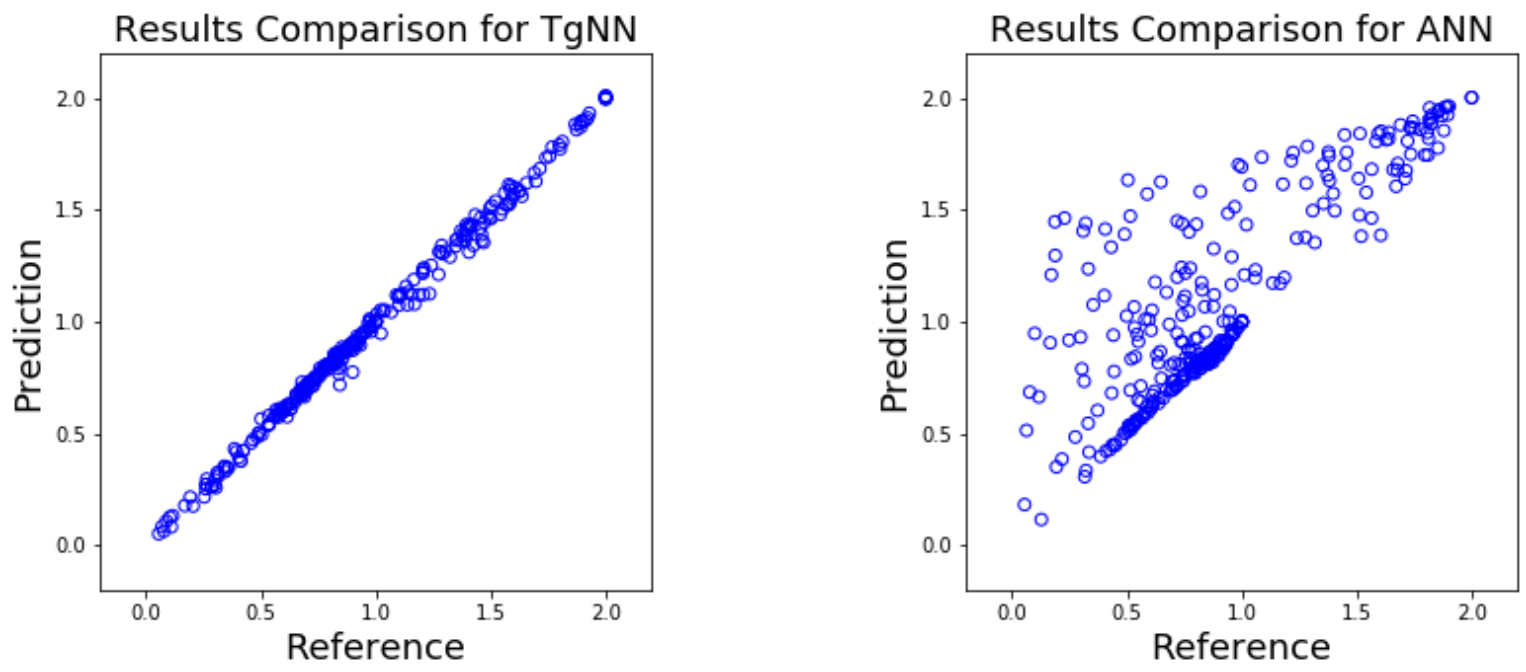

**Figure 7.** Correlation between the reference and predicted hydraulic head with changed boundary conditions.

**Table 3.** Relative $L_2$ error and $R^2$ score of TgNN and ANN.

| | **TgNN** | **ANN** |
|---|---|---|
| **Relative $L_2$ error** | 2.605880e-02 | 3.070765e-01 |
| **$R^2$ score** | 9.962868e-01 | 4.843825e-01 |

### 3.3 Predicting the future response in the presence of data noise and outliers

In the previous cases, the training data are clean, i.e., noiseless. However, in practice, observation data from monitors or sensors may be degraded by noise, and outliers may exist in the data when monitors are out of order. Therefore, in this subsection, we will test the performance of TgNN when noise and outliers exist.

#### 3.3.1 *Predicting the future response from noisy data*

In this work, noise is added into the observation data in the following manner:

$$h(t,x,y) = h(t,x,y) + h_{diff}(x,y) \times a\% \times \varepsilon \quad (33)$$

where $h_{diff}(x,y)$ denotes the maximum of the response difference at monitoring location ($x$, $y$) during the monitoring process; $a$ is a percentage; and $\varepsilon$ denotes the uniform random variable ranging from -1 to 1. **Figure 8** shows the observed hydraulic head at $t$=3 and $y$=320 with 0%, 5%, 10%, and 20% noise, respectively.

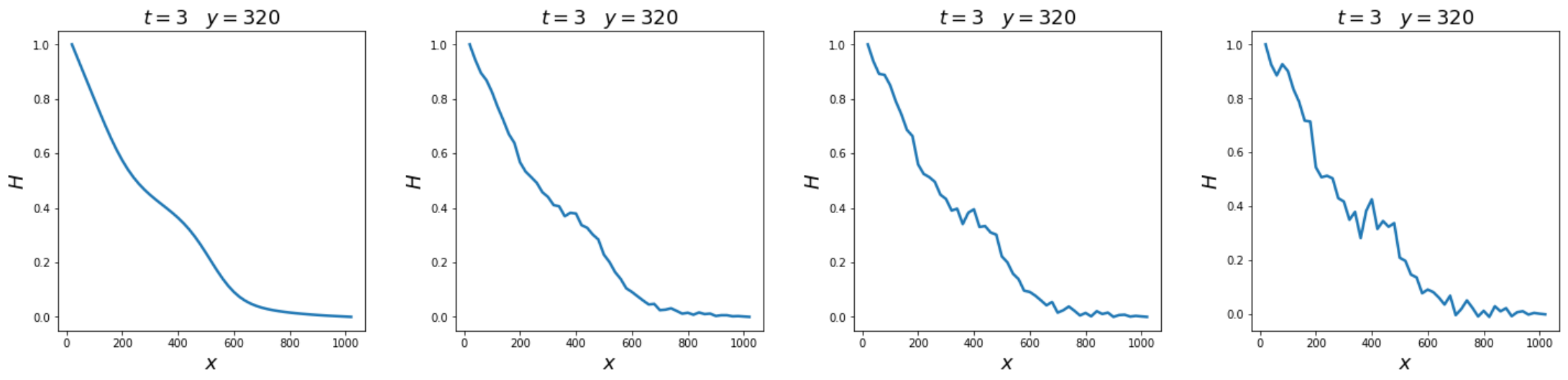

**Figure 8.** Observation data at $t$=3 and $y$=320 with 0%, 5%, 10%, and 20% noise.

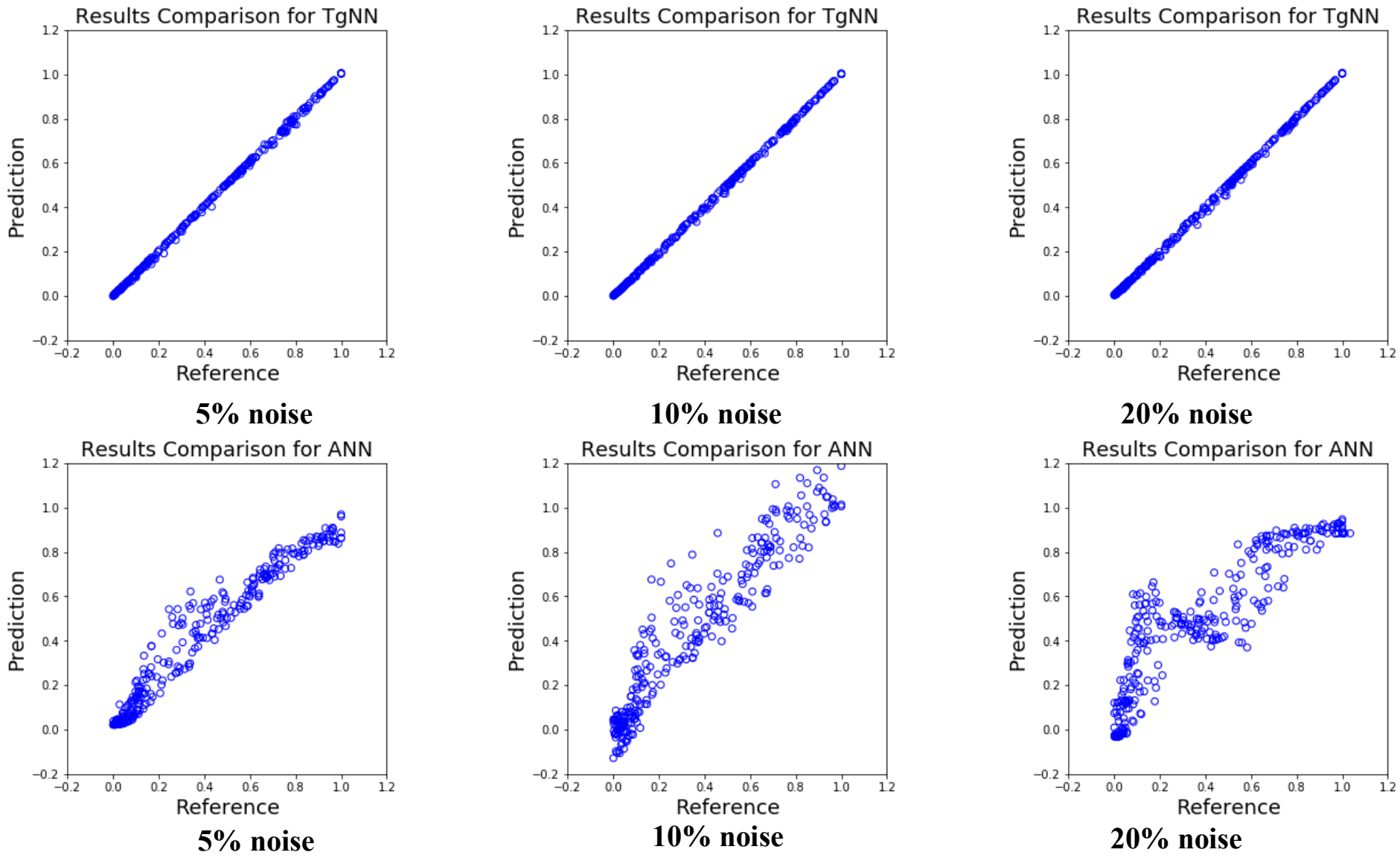

**Figure 9.** Correlation between the reference and predicted hydraulic head when noise exists.

**Table 4.** Relative $L_2$ error and $R^2$ score of TgNN and ANN when noise exists.

| | **Relative $L_2$ error** | | **$R^2$ score** | |
|---|---|---|---|---|
| | **TgNN** | **ANN** | **TgNN** | **ANN** |
| **5% noise** | 1.489959e-02 | 1.359368e-01 | 9.994565e-01 | 9.658794e-01 |
| **10% noise** | 1.001120e-02 | 3.113413e-01 | 9.997546e-01 | 8.405199e-01 |
| **20% noise** | 1.353335e-02 | 3.282400e-01 | 9.995516e-01 | 8.209727e-01 |

The predictions of the TgNN model and the ANN model trained from noisy data are shown in **Appendix B.3**. It is found that the predictions of TgNN are almost unaffected by the noise, while the ANN model is seriously deteriorated and produces worse predictions, which indicates that TgNN possesses superior robustness to ANN. The correlation between the reference and predicted hydraulic head, shown in **Figure 9**, and the relative $L_2$ error and $R^2$ score, shown in **Table 4**, provide similar observations.

### 3.3.2 *Predicting future response in the presence of outliers*

When monitors or sensors are out of order, some totally unreasonable data may be present in the observation, for example, in this case, the hydraulic head higher than $1[L]$ or lower than $0[L]$. In order to test the influence of outliers, here 5%, 7%, and 10% of observation data are set to be outliers, respectively, and their values are randomly generated from a uniform distribution ranging from 1 to 2.

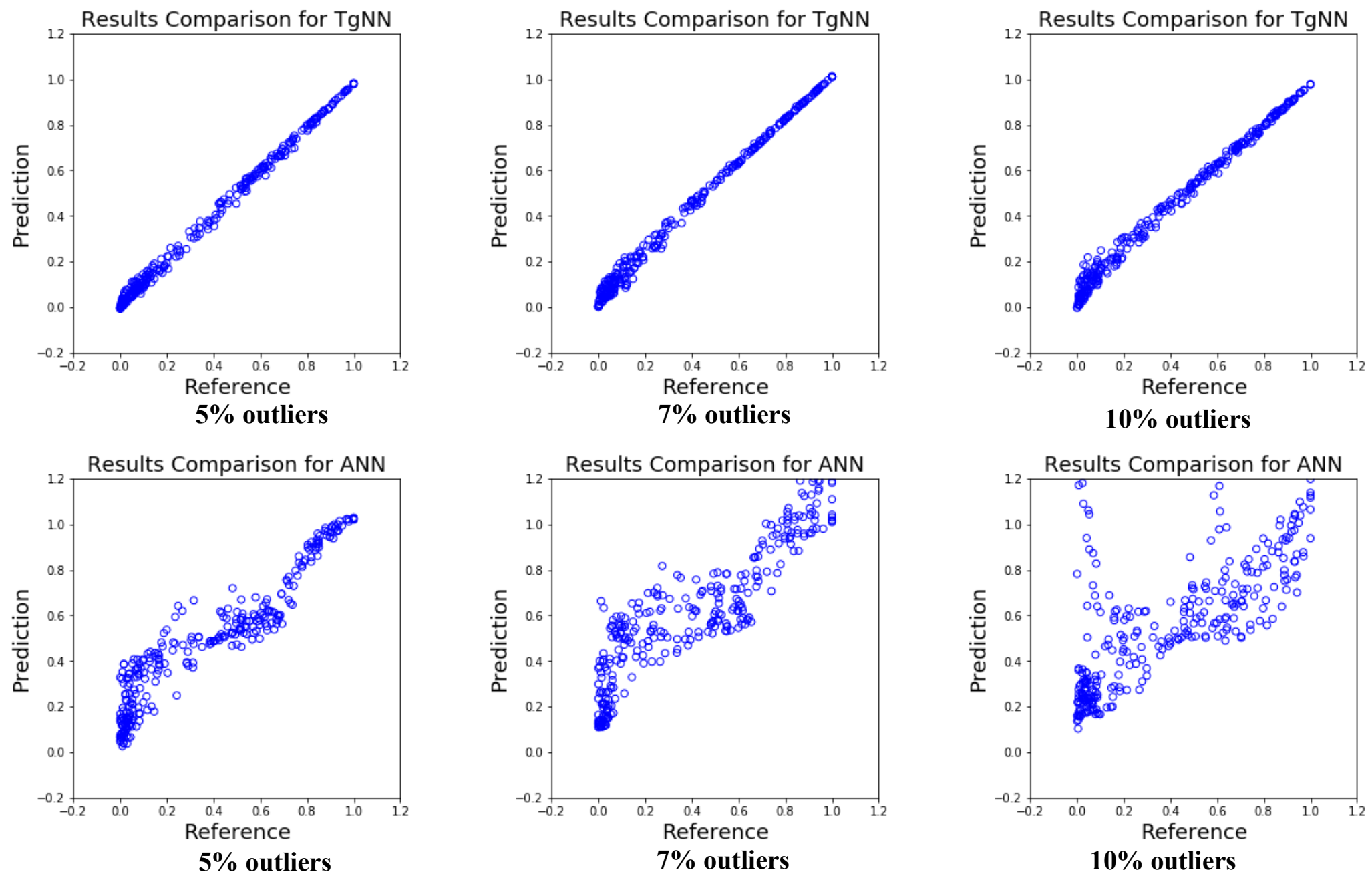

**Figure 10.** Correlation between the reference and predicted hydraulic head in the presence of outliers.

The prediction results at time step 30 and 50 with different amounts of outliers are presented in **Appendix B.3**, and the correlation between the prediction and reference data is shown in **Figure 10**. It can be seen that, as outliers increase, the accuracy of the model becomes worse, especially in the early time (prediction at time step 30). However, as the simulation time goes on, the effect of outliers is reduced by the embedded scientific knowledge in the model (prediction at time step 50). It can be found that the TgNN can effectively resist the effect of outliers by introducing the physical and engineering constraints in the training process. The PDE and expert knowledge (e.g., **Eq. (29)**) terms will introduce penalty to the loss function when they are violated, which avoids the model being misguided by the outliers. On the other hand, the ANN, which highly depends on the data, will be seriously degraded by outliers and produce incorrect predictions.

**Table 5.** Relative $L_2$ error and $R^2$ score of TgNN and ANN when outliers exist.

| | **Relative $L_2$ error** | | **$R^2$ score** | |
|---|---|---|---|---|
| | **TgNN** | **ANN** | **TgNN** | **ANN** |
| **5% outliers** | 4.337816e-02 | 3.107828e-01 | 9.953933e-01 | 8.292832e-01 |
| **7% outliers** | 8.190370e-02 | 5.363573e-01 | 9.835769e-01 | 5.135411e-01 |
| **10% outliers** | 1.020689e-01 | 6.252584e-01 | 9.744943e-01 | 2.743178e-01 |

### 3.4 Transfer learning based on TgNN

In this subsection, the concept of 'transfer learning' is introduced to deal with the 'boundary condition changing' situation in a more efficient manner (Pan & Yang, 2009). Unlike the situation described in subsection 3.2, in which the changing of boundary condition is known in advance and the information is already incorporated into the loss function, here it is assumed that the new boundary condition is known after the model training. In order to avoid retraining the whole model, transfer learning constitutes an appropriate technique.

In this case, the hydraulic head at the end $x = 1020[L]$ rises to $2[L]$ from $0[L]$ at time step 20, and the TgNN model has been trained using the data from the first 20 time steps. In order to predict the following hydraulic head based on the trained TgNN model, transfer learning is adopted. As discussed in Sun et al. (2019), the shallow layers of the network extract information about a particular system, while the deeper layers process the extracted information. Therefore, for performing transfer learning in this case, the weights and bias of the last four layers are fixed, which have already learned the information of the global governing equation. In addition, only the parameters of the first three layers are retrained to extract the information of the new boundary conditions. During the retraining process, it is started with the parameters of the pre-trained model. Using transfer learning, the computational cost for training the new TgNN model with new boundary condition is decreased.

In the transfer learning step, no observation data are needed, and the loss function can be expressed as:

$$L(\theta) = \lambda_{PDE} MSE_{PDE} + \lambda_{BC} MSE_{BC} + \lambda_{IC} MSE_{IC} + \lambda_{EC} MSE_{EC} + \lambda_{EK} MSE_{EK} \tag{34}$$

where the initial condition can be obtained from the prediction of the pre-trained model at time step 20.

In order to illustrate the effectiveness and efficiency of the transfer learning mode, two contrasting examples are introduced, following the work by Sun et al. (2019): retrain the first three layers while the last four layers are randomly initialized, and retrain the network completely.

**Table 6** and **Figure 11** show the prediction results of the transfer learning model and the two contrasting examples. Compared with contrasting example 1, whose network parameters of the last four layers are randomly initialized and fixed during the training process, transfer learning model has better accuracy, which may indicate that the pre-trained parameters of the last four layers learned some information of the system during the former training process, and the information is transferable when dealing with similar new systems (Sun et al., 2019). Compared with contrasting example 2, whose parameters are retrained totally, the transfer learning model only exhibits a slight advantage regarding accuracy, while efficiency is noticeably improved since fewer parameters need to be trained. For this case, training time is reduced by 16.7%, and this efficiency improvement can be larger when dealing with more complicated problems. From this case, it can be seen that the transfer learning framework can be integrated into the TgNN model for handling variable condition situations.

**Table 6.** Transfer learning experiments results.

| | **First three layers** | **Last four layers** | **Relative $L_2$ error** | **$R^2$ score** | **Training time/s** |
|---|---|---|---|---|---|
| **Transfer learning** | Pre-trained/trainable | Pre-trained/fixed | 2.2941e-02 | 9.9798e-01 | 755.40 |
| **Contrasting example 1** | Randomly initialized/trainable | Randomly initialized/fixed | 1.5598e-01 | 9.0668e-01 | 758.06 |
| **Contrasting example 2** | Randomly initialized/trainable | Randomly initialized/trainable | 2.9222e-02 | 9.9672e-01 | 907.00 |

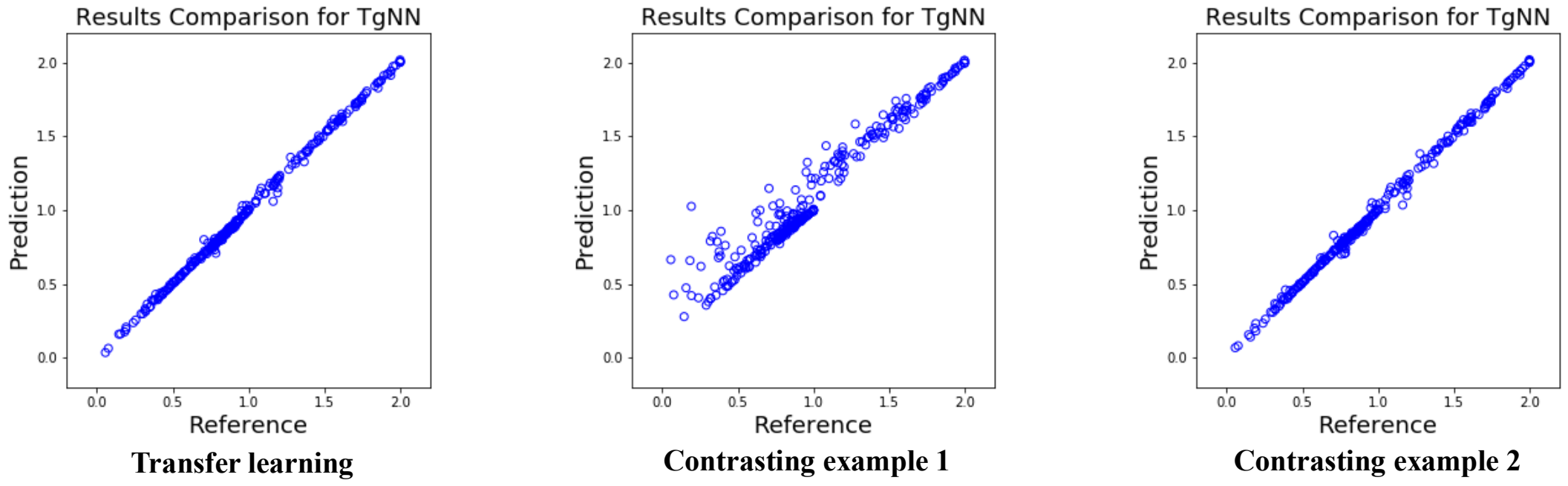

**Figure 11.** Correlation between the reference and predicted hydraulic head for different model settings.

### 3.5 Incorporating engineering control

In this subsection, in order to assess the effect of engineering control (EC) in the TgNN framework, another case is introduced, which contains a sink term in the governing equation. The domain size and the physical parameters are the same as in previous cases, and the difference is that the boundary conditions are prescribed for the left and the right sides with prescribed hydraulic heads of $202[L]$ and $198[L]$, respectively. In addition, a pumping well is located at $x$=$520[L]$ and $y$=$520[L]$, and the volumetric pumping rate of the well is Q=50 $[L^3/T]$. The engineering control of the case is that the hydraulic head must not be lower than 81 $[L]$ (which can be from the hydraulic head drawdown control of local government policy). When the hydraulic head at the well location reaches or is close to the prescribed value, the well must be operated in a different manner (with reduced pumping rates or at a prescribed bottom well pressure) in order not to violate the engineering control. Although, in the reference case, the well is switched to the prescribed pressure when the drawdown reaches the preset value, this exact operation schedule may not be publicly accessible. Therefore, it can be challenging for the TgNN to cope with this problem via changing the well conditions. However, this problem can be approached alternatively through incorporating engineering control into the TgNN model.

Similar to section 3.1, it is assumed that the hydraulic head distribution at the first 20 time steps are monitored, and 2000 data points are extracted at each time step as training data. Moreover, the hydraulic head at the well point is monitored for the first 20 time steps. The well pumping rate is assumed to be constant throughout the entire period. The governing equation at the well point is:

$$S_s \frac{\partial h}{\partial t} = \frac{\partial}{\partial x}\left( K(x,y)\frac{\partial h}{\partial x} \right) + \frac{\partial}{\partial y}\left( K(x,y)\frac{\partial h}{\partial y} \right) - \frac{Q}{\Delta x \Delta y} \tag{35}$$

So, this partial differential equation residual and loss can be represented as follows:

$$f_{well} := S_s \frac{\partial N_h(t,x,y;\theta)}{\partial t} - K\frac{\partial^2 N_h(t,x,y;\theta)}{\partial x^2} - K\frac{\partial^2 N_h(t,x,y;\theta)}{\partial y^2} + \frac{Q}{\Delta x \Delta y} \tag{36}$$

$$MSE_{PDE-well} = \frac{1}{N_{well}} \sum_{i=1}^{N_{well}} \left| f(t_{well}^i, x_{well}^i, y_{well}^i) \right|^2 \tag{37}$$

In addition, the engineering control term can be expressed as:

$$MSE_{EC} = \frac{1}{N_{EC}} \left( \sum_{i=1}^{N_{EC}} \left| \mathrm{ReLU}(H_{EC} - N_h(t,x,y;\theta)) \right|^2 \right) \tag{38}$$

where $H_{EC}$=81 $[L]$ in this case. Then, the loss function in this case is:

$$\begin{aligned} L(\theta) = {} & \lambda_{PDE} MSE_{PDE} + \lambda_{PDE-well} MSE_{PDE-well} + \lambda_{DATA} MSE_{DATA} \\ & + \lambda_{BC} MSE_{BC} + \lambda_{IC} MSE_{IC} + \lambda_{EC} MSE_{EC} \end{aligned} \tag{39}$$

When the engineering control (a certain drawdown in this case) is violated, as shown in **Eq. (38)**, a penalty occurs. Then, **Eq. (36)** with a constant pumping rate is no longer strictly valid. This is done via the reduction of its weight in the loss function of **Eq. (39)**.

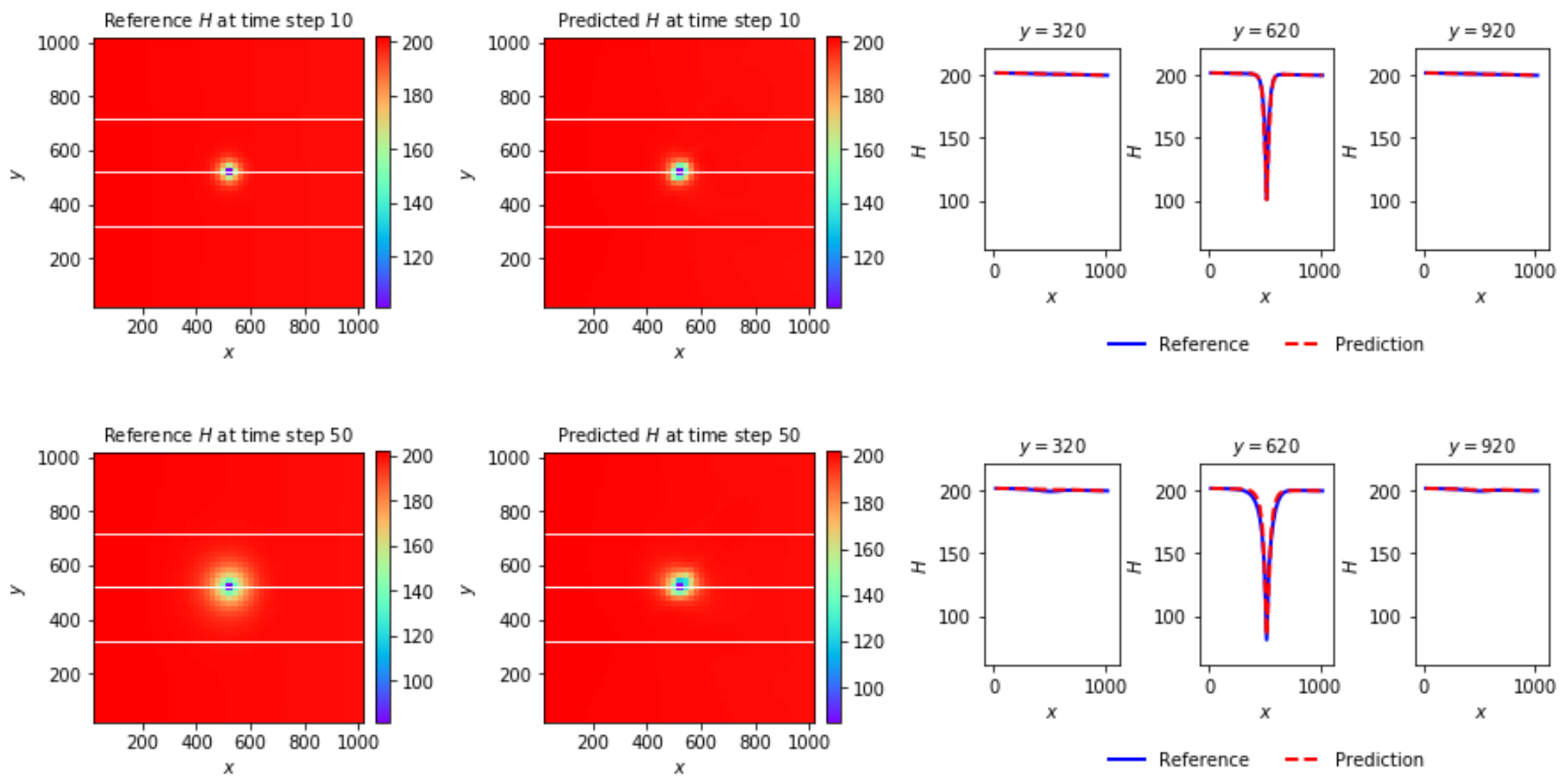

**Figure 12.** Prediction results of TgNN with engineering control.

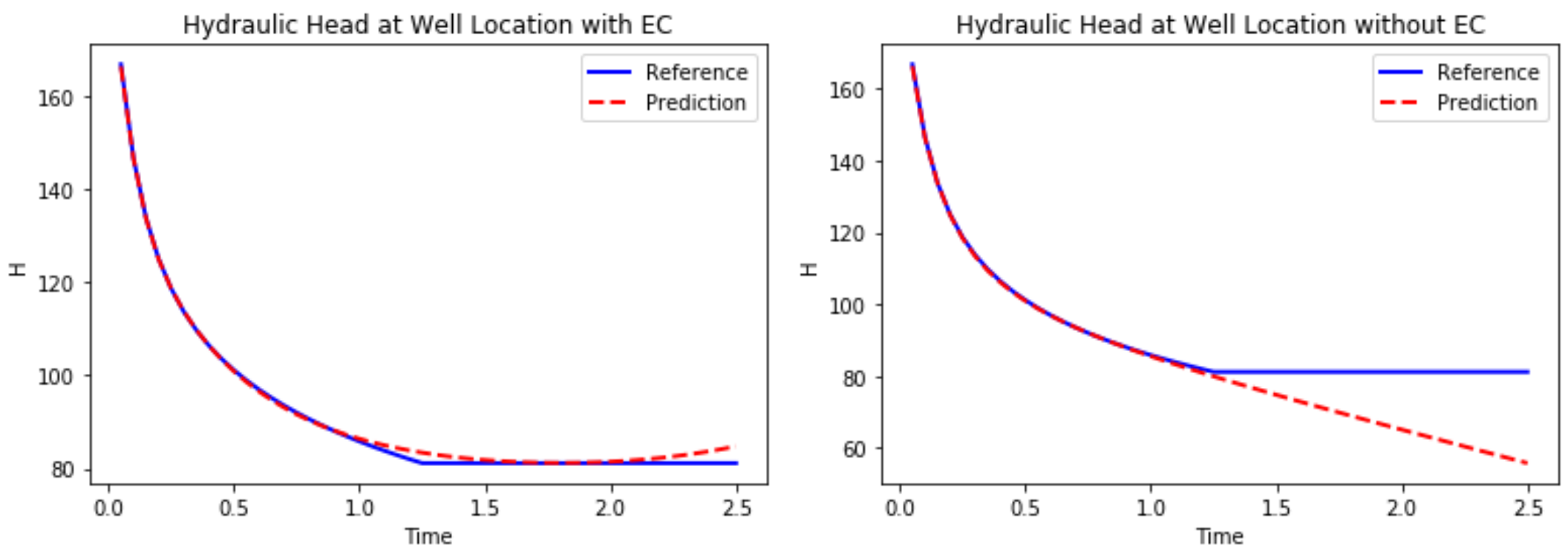

**Figure 13.** Prediction results at the well location of TgNN with and without engineering control.

The prediction results at time step 10 and 50 of this case are shown in **Figure 12.** It can be observed that the TgNN model can predict the hydraulic head distribution well for this case. To discern the effect of engineering control, we compare the prediction of hydraulic head at the well location from TgNN with and without engineering control, as shown in **Figure 13**. The prediction results of TgNN without engineering control are presented in **Appendix C**. It can be seen that the TgNN model can make a similar prediction to the constant pressure pumping situation by applying EC in the loss function. However, the hydraulic head value can go much lower than the control value $H_{EC}$ when there is no EC term in the loss function. From this case, the necessity of the EC term in the TgNN framework for solving some cases with practical engineering operations can be seen, which cannot be accomplished by incorporating only the PDE constraint.

## 4. Discussions and Conclusions

In this work, we proposed the Theory-guided Neural Network (TgNN), a novel framework for deep learning of subsurface flow, which incorporates scientific knowledge, such as governing equations, boundary conditions, and practical experience (such as expert knowledge and engineering controls) into the model. Through TgNN, deep learning is not only driven by data, as that in the Deep Neural Network (DNN), but also by physical laws and engineering theory, which can assist the model to achieve better accuracy, generalization ability, and robustness.

Subsurface porous medium is usually heterogeneous in practice, which makes the hydraulic conductivity in the governing equation not a constant coefficient, but a heterogeneous field. In this work, Karhunen–Loeve expansion (KLE) is introduced to parameterize the heterogeneous field, which enables TgNN to deal with problems with heterogeneous model parameters more effectively.

The applicability and performance of the proposed TgNN framework are tested by several cases with different situations: predicting future response, predicting with changed boundary conditions, training from noisy data or outliers, transfer learning, and incorporating engineering controls. Compared with ANN, TgNN achieves results that are far superior in these scenarios.

The TgNN constitutes a type of data-driven approach. In this work, some (early time) data are assumed to be given either from model simulations or field measurements, and the parameters of the heterogeneous field are assumed to be known. On the one hand, the intent of the TgNN is not to complete with numerical solutions of the governing equations (i.e., model-driven approaches). Instead of solving the flow equations numerically, the TgNN learns from the available data in an efficient manner, while simultaneously adhering to the known physical principles and engineering constraints. Future studies may address such issues as computational efficiency compared to pure numerical solutions, especially in the case of large scale models (either with a large number of grid blocks or multiple, coupled processes), as well as the impacts of available data, and the quantity and quality of model parameters (e.g., hydraulic conductivity). On the other hand, the proposed TgNN framework can serve as a foundation for performing certain other tasks. For example, TgNN can be utilized as a surrogate model for performing efficient uncertainty quantification and parameter inversion, which can produce physically reasonable and consistent results with high accuracy.

## Acknowledgements

This work is partially funded by the National Natural Science Foundation of China (Grant No. U1663208 and 51520105005) and the National Science and Technology Major Project of China (Grant No. 2017ZX05009-005 and 2017ZX05049-003). The

data used in this paper are made available for download through the following link: https://figshare.com/articles/Data_for_TgNN/10032980/1(doi:10.6084/m9.figshare.10032980)

## Appendix A

This appendix provides analytical or semi-analytical solutions of eigenvalues and eigenfunctions for the separable exponential covariance function.

For a one-dimensional stochastic process with an exponential covariance function $C_Z(x,x') = \sigma_Z^2 \exp(-|x-x'|/\eta)$, where $\sigma_Z^2$ and $\eta$ are the variance and the correlation length of the random field, respectively, the eigenvalues and eigenfunctions can be solved analytically or semi-analytically, as given below (Zhang & Lu, 2004):

$$\lambda_i = \frac{2\eta\sigma_Z^2}{\eta^2\omega_i^2 + 1} \tag{A.1}$$

$$f_i(x) = \frac{1}{\sqrt{(\eta^2\omega_i^2 + 1)L/2 + \eta}} \left[\eta\omega_i \cos(\omega_i x) + \sin(\omega_i x)\right] \tag{A.2}$$

where $\omega_i$ are positive roots of the following characteristic equation:

$$(\eta^2\omega^2 - 1)\sin(\omega L) = 2\eta\omega\cos(\omega L) \tag{A.3}$$

For a two-dimensional case, if we assume that the covariance function is separable, i.e., $C_Z(\mathbf{x},\mathbf{x'}) = \sigma_Z^2 \exp(-|x_1 - x_2|/\eta_x - |y_1 - y_2|/\eta_y)$, the eigenvalues and eigenfunctions can be obtained by combining them in each direction, as given below ($x$ direction) (Zhang & Lu, 2004):

$$\lambda_i^{(x)} = \frac{2\eta_x\sigma_Z^2}{\eta_x^2\left(\omega_i^{(x)}\right)^2 + 1} \tag{A.4}$$

$$f_i^{(x)}(x) = \frac{1}{\sqrt{\left(\eta_x^2\left(\omega_i^{(x)}\right)^2 + 1\right)L_x/2 + \eta_x}} \left[\eta_x\omega_i^{(x)} \cos(\omega_i^{(x)} x) + \sin(\omega_i^{(x)} x)\right] \tag{A.5}$$

where $\omega_i$ are positive roots of the following characteristic equation:

$$(\eta_x^2\omega^2 - 1)\sin(\omega L_x) = 2\eta_x\omega\cos(\omega L_x) \tag{A.6}$$

The eigenvalues and eigenfunctions are then combined in each direction, which can be expressed as follows:

$$\lambda_i = \frac{1}{\sigma_Z^2}\lambda_j^{(x)}\lambda_k^{(y)} \tag{A.7}$$

$$f_i(x,y) = f_j^{(x)}(x) f_k^{(y)}(y) \tag{A.8}$$

## Appendix B

This Appendix provides detailed prediction results in different scenarios for the case without a sink term.

### B.1 Prediction with changed boundary conditions

In this case, it is assumed that the hydraulic head distribution at the first 18 time steps is monitored, and 1000 data points are extracted at each time step as training data. The hydraulic head distribution at the following 32 time steps are predicted. The predictions from TgNN and ANN at time step 50 (the last step) with hydraulic conductivity field (b), (c), and (d) are shown in **Figure B.1**. It can be seen that the predictions of TgNN match the reference values well and are superior to the predictions of ANN.

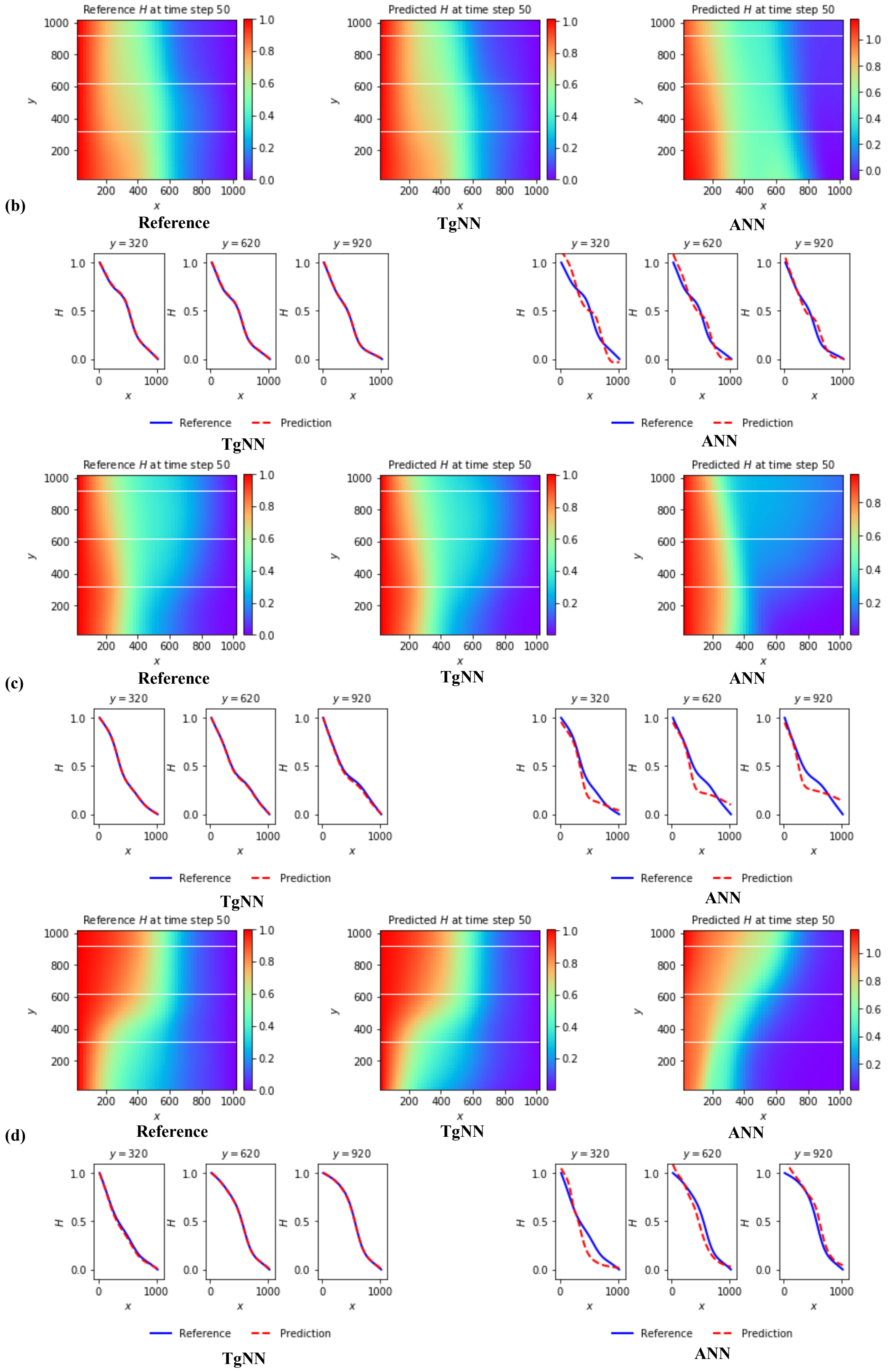

**Figure B.1.** Prediction results of TgNN and ANN for hydraulic conductivity field (b), (c), and (d).

## B.2 Prediction with changed boundary conditions

In this case, the prescribed hydraulic head at the end $x = 1020\,[L]$ rises to $2\,[L]$ from $0\,[L]$ at time step 20. The hydraulic heads at the first 20 time steps are monitored, 1000 data points are selected at each time step as training data, and no data exist after the boundary condition changes. The prediction results of the hydraulic head at time step 30, 40, and 50 from TgNN and ANN when changing the boundary conditions are shown in **Figure B.2**. It can be observed that the TgNN obtains better performance than the ANN.

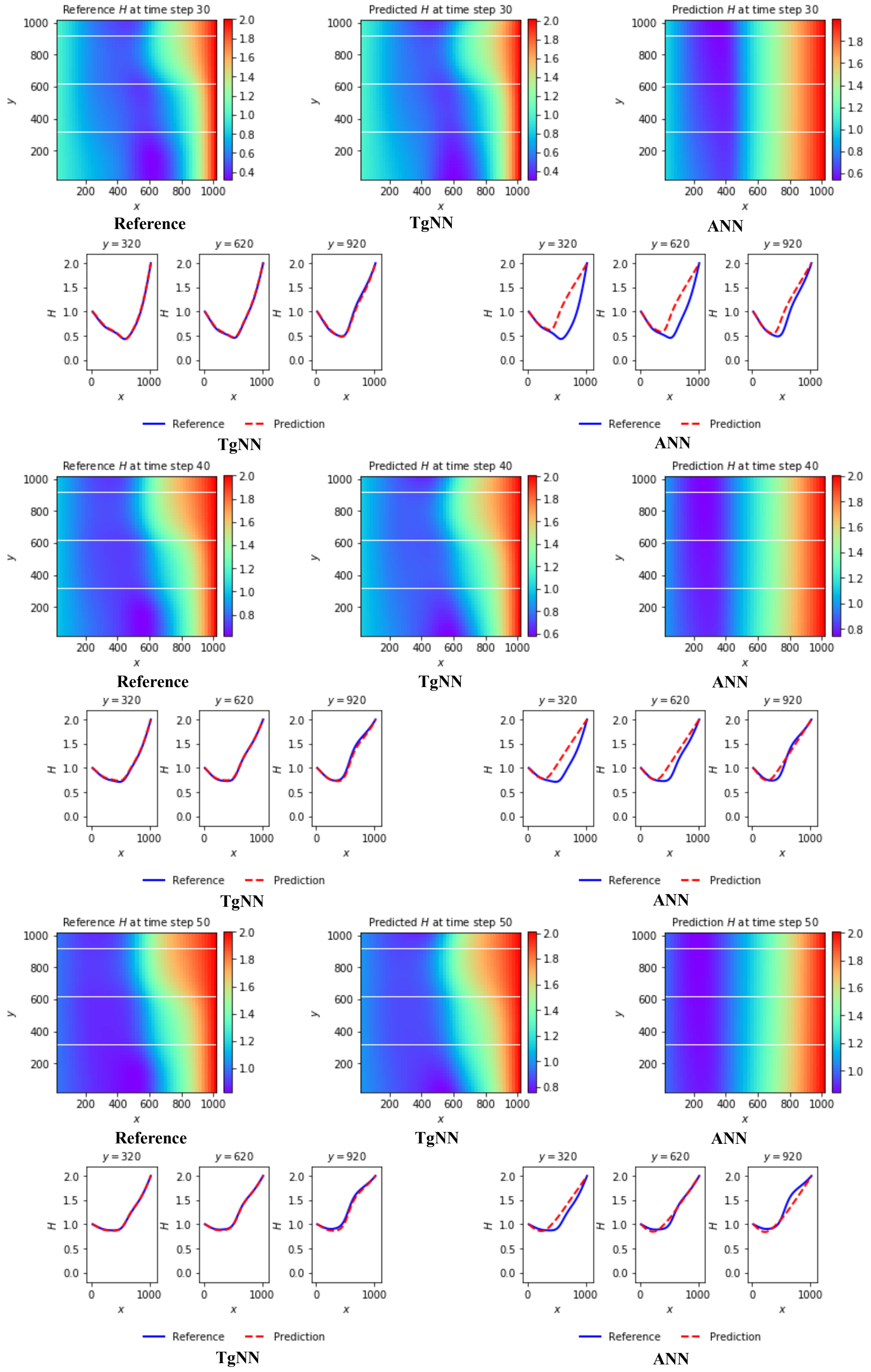

**Figure B.2.** Predictions of hydraulic head at different time step (30, 40, and 50) via TgNN and ANN when changing the boundary conditions.

## B.3 Prediction of future response in the presence of data noise and outliers

In this case, noise and outliers are added into the monitored training data, and different levels of noise (5%, 10%, and 20%) and outliers (5%, 7%, and 10%) are considered. The prediction results at time step 50 of the TgNN model and the ANN model trained from different amounts of data noise are presented in **Figure B.3(a).** The results demonstrate that the TgNN model has better robustness than the ANN model when noise exists in the data.

The prediction results of the TgNN model and the ANN model at time step 30 and 50 with different amounts of outliers are shown in **Figure B.3(b).** The prediction at time step 30 is not as good as that at time step 50, which is because the effect of outliers is reduced by the incorporated scientific knowledge in the model as time passes. TgNN achieves better results than ANN.

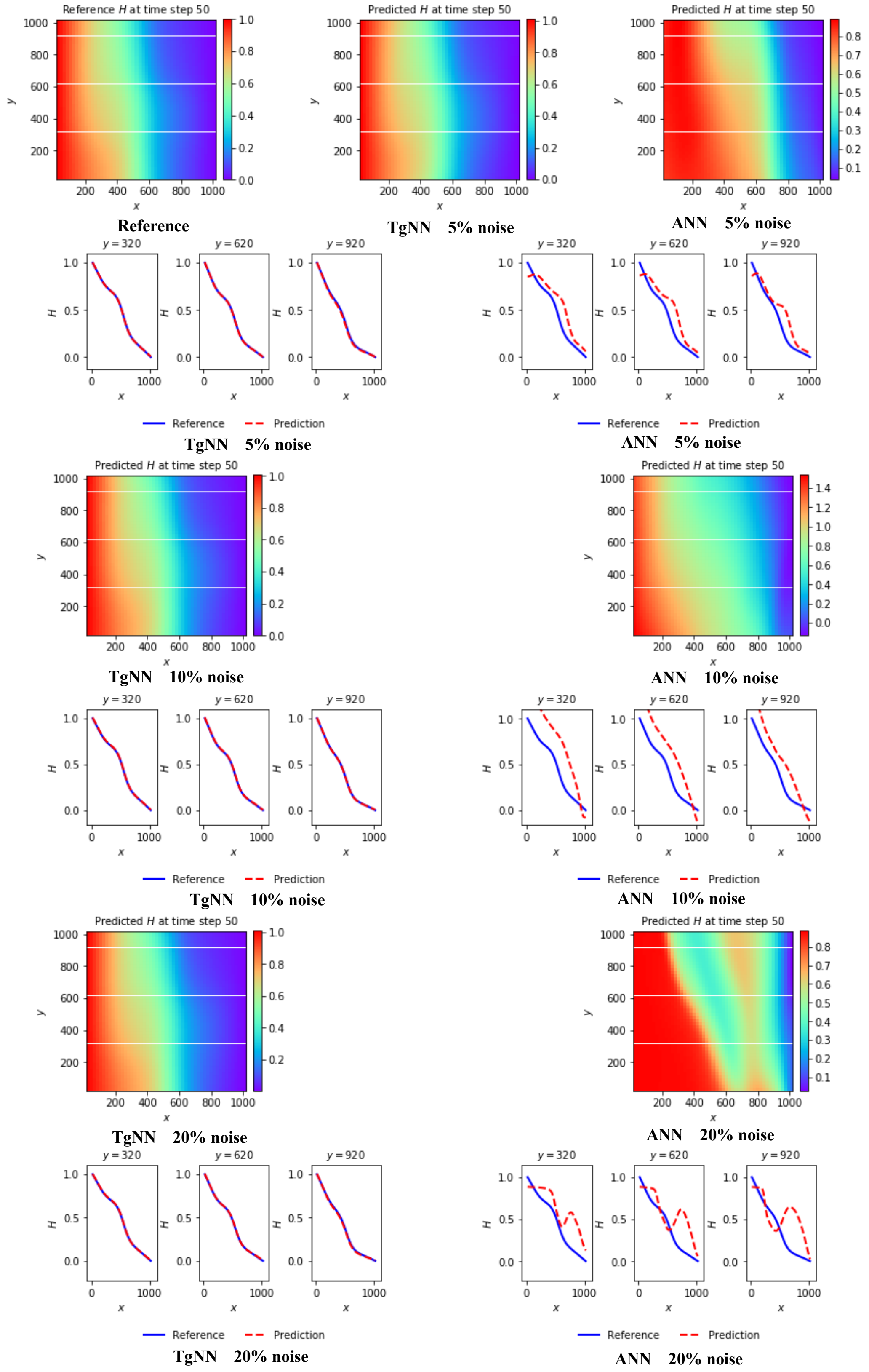

**Figure B.3(a).** Predictions of the TgNN model and the ANN model trained from noisy data.

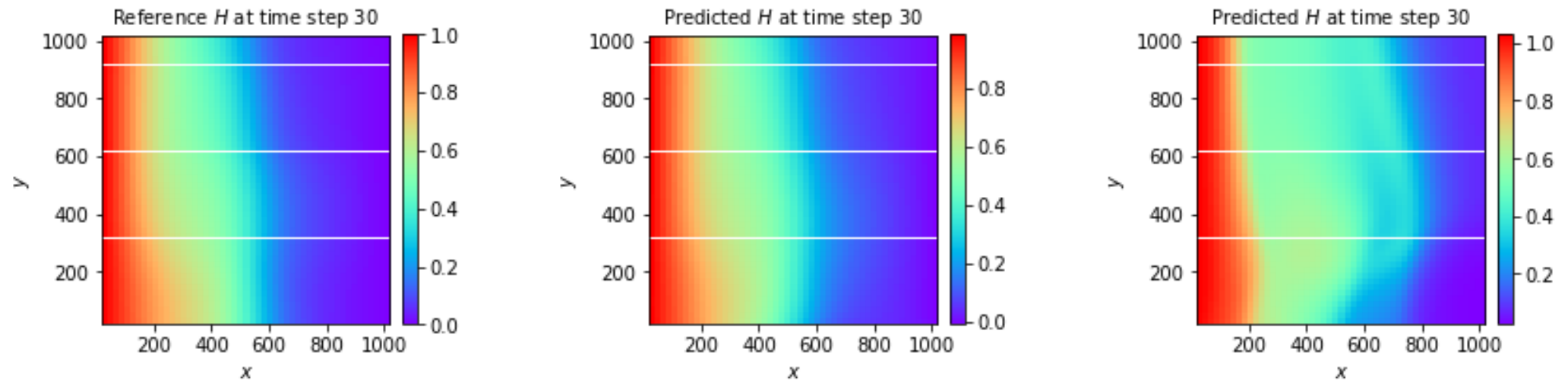

Reference TgNN 5% outliers ANN 10% outliers

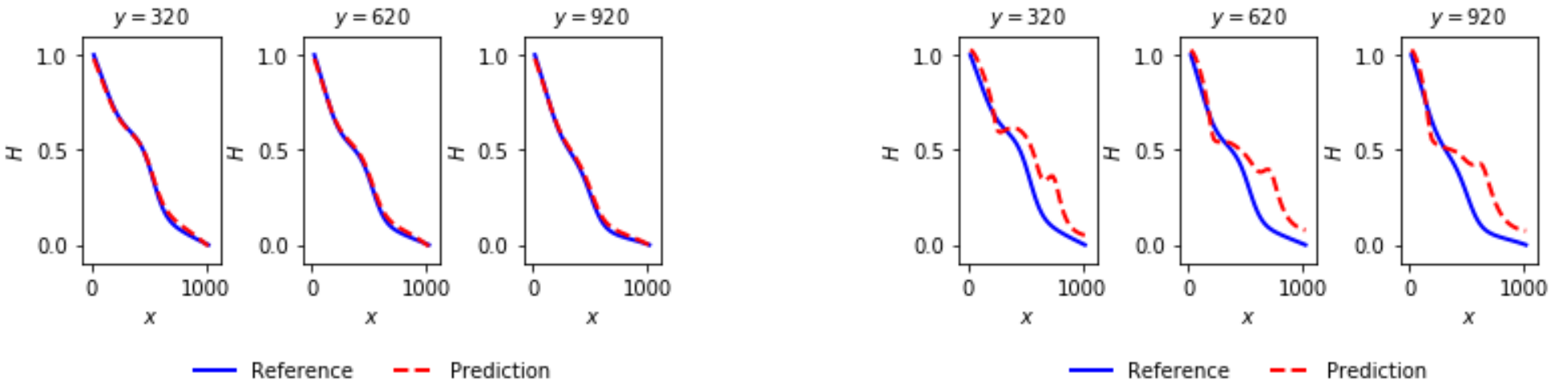

TgNN predictions at time step 30 with 5% outliers ANN predictions at time step 30 with 5% outliers

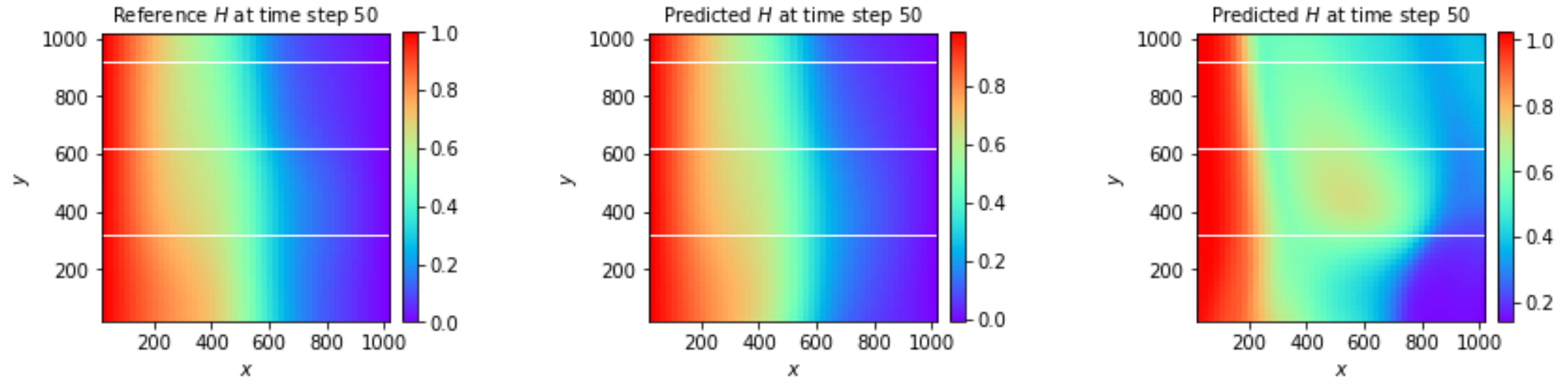

Reference TgNN 5% outliers ANN 5% outliers

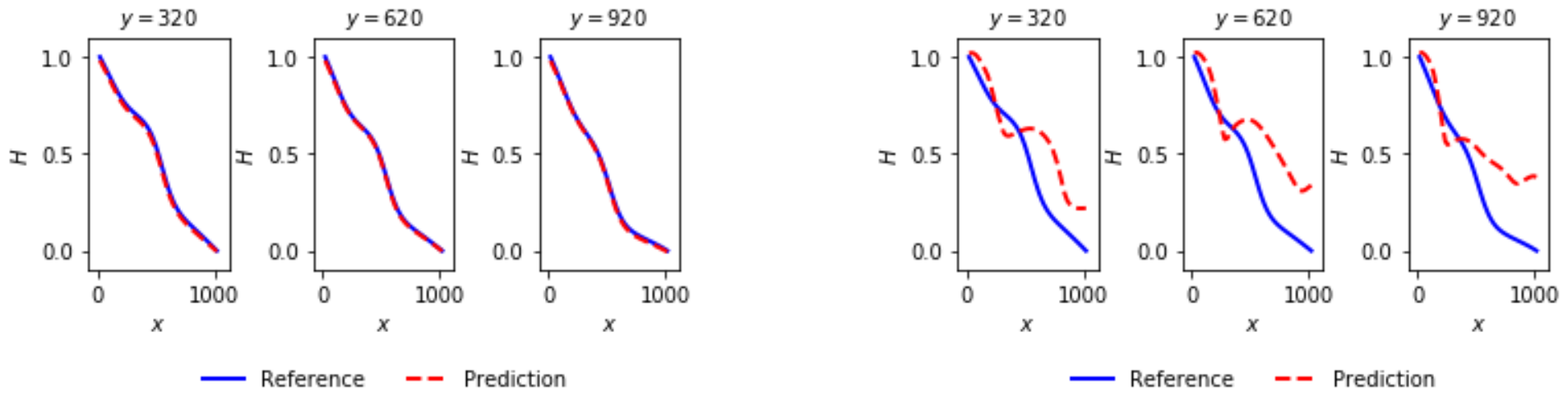

TgNN predictions at time step 50 with 5% outliers ANN predictions at time step 50 with 5% outliers

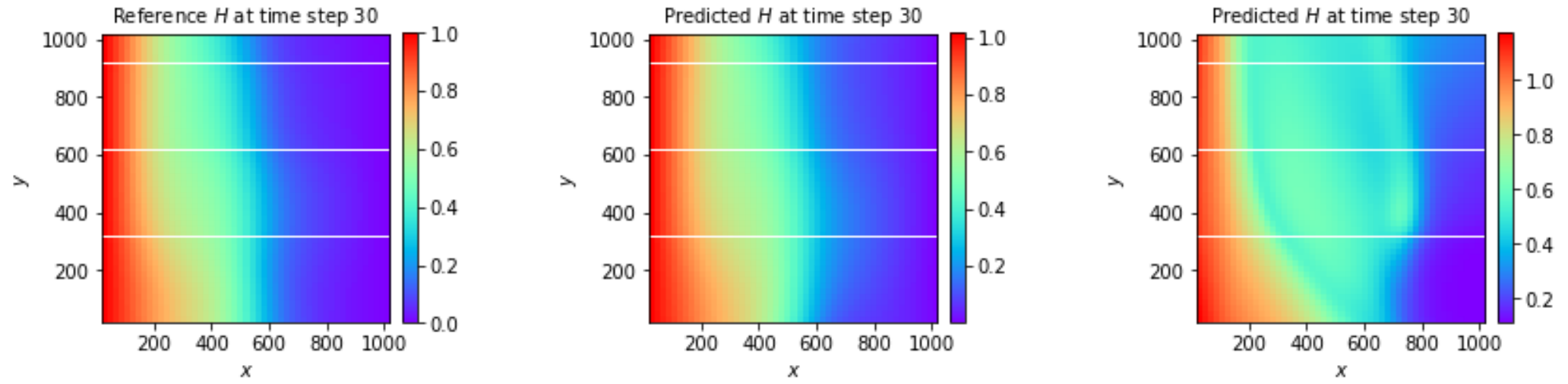

Reference TgNN 7% outliers ANN 10% outliers

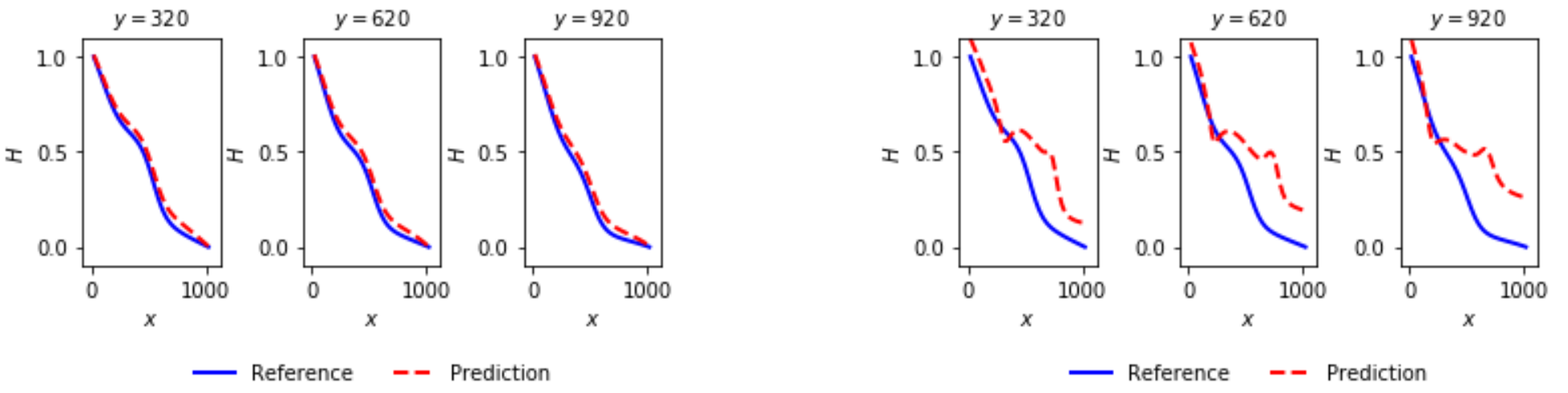

TgNN predictions at time step 30 with 7% outliers ANN predictions at time step 30 with 7% outliers

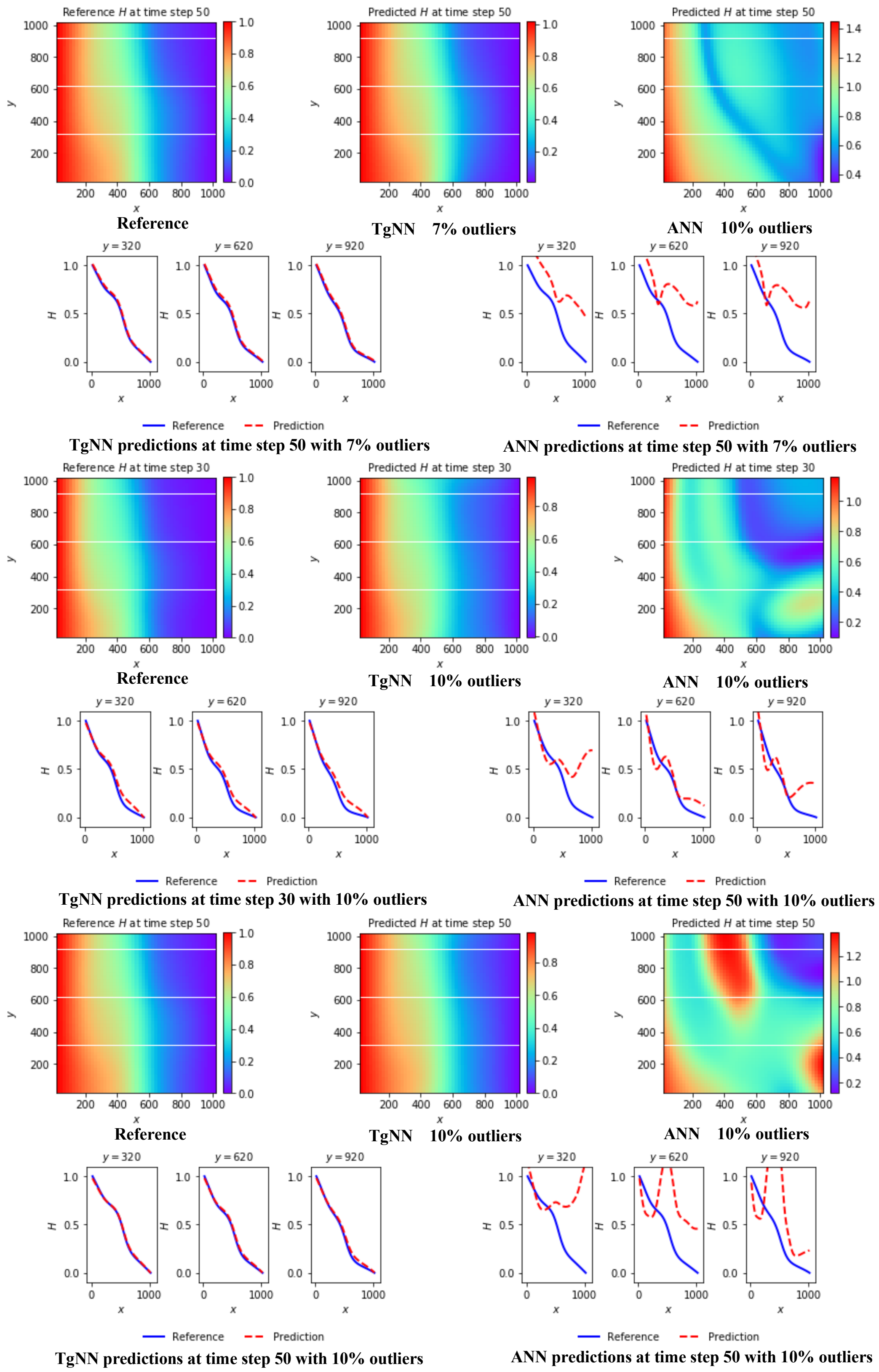

**Figure B.3(b).** Predictions of the TgNN model and the ANN model at time step 30 and 50 in the presence of outliers in the training data.

## Appendix C

This Appendix provides detailed prediction results for the case with a sink term and well operation controls. In this case, a subsurface flow with a pumping well is considered. The pumping well is located at $x$=520 $[L]$ and $y$=520 $[L]$, and the volumetric pumping rate of the well is Q=50$[L^3/T]$. However, when the hydraulic head at the well location reaches the hydraulic head control value 81$[L]$, the well will be operated to maintain the prescribed hydraulic head in order to avoid excessive pressure drawdown in the well. The predictions of hydraulic head at time step 20, 30, 40, and 50 from TgNN with and without engineering control are shown in **Figure C.1.** It can be seen that the hydraulic head at the well location will exceed the control value when there is no engineering control in the model.

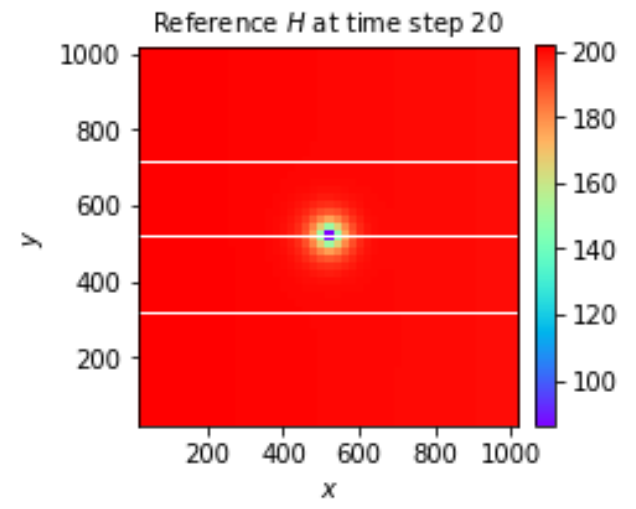

Reference

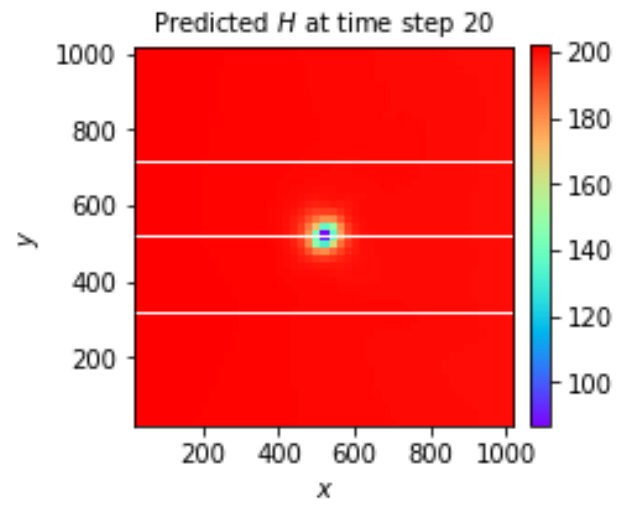

TgNN with EC

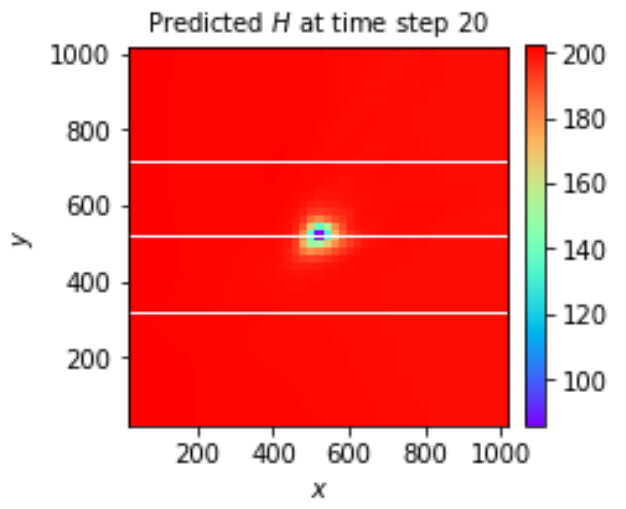

TgNN without EC

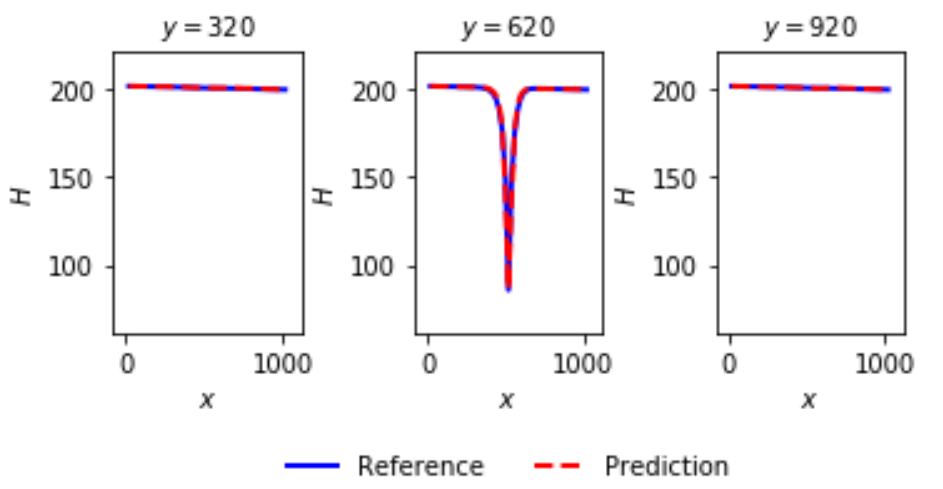

TgNN with EC

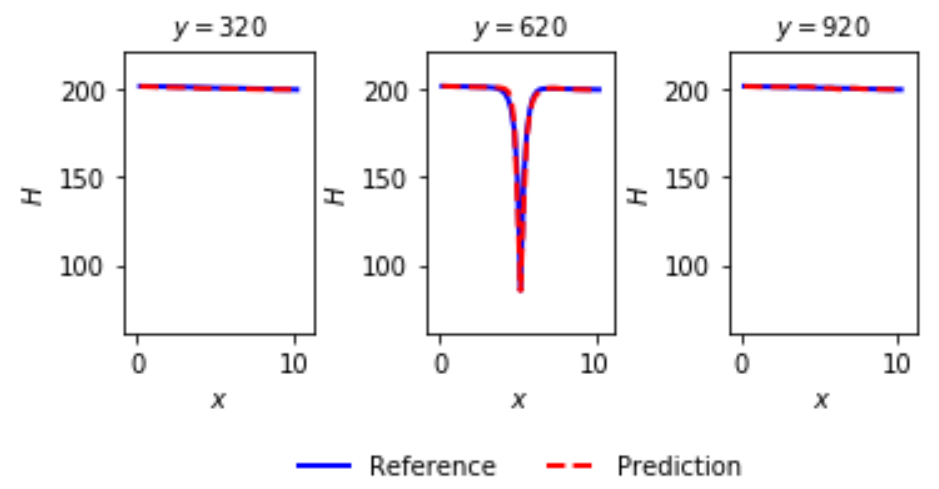

TgNN without EC

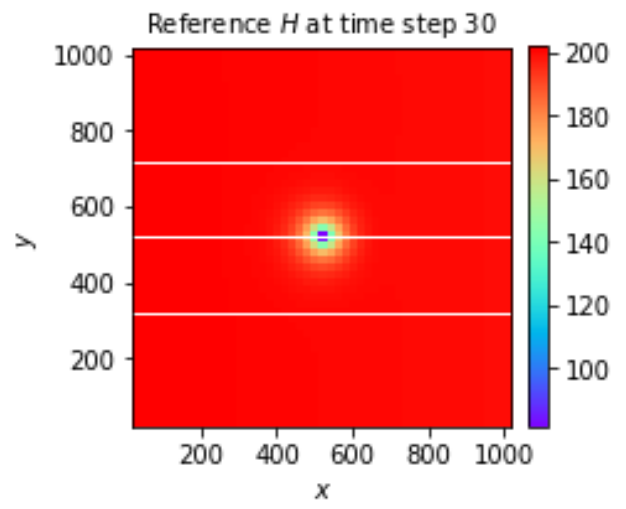

Reference

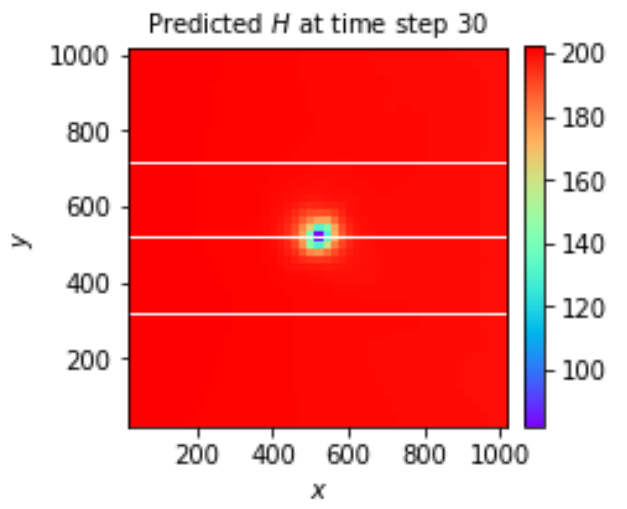

TgNN with EC

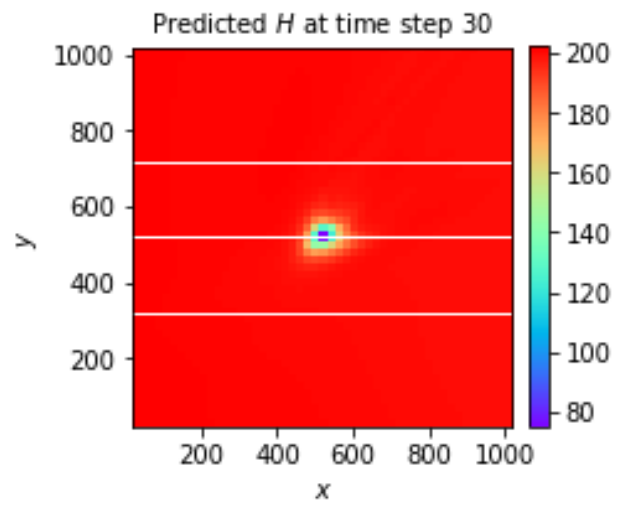

TgNN without EC

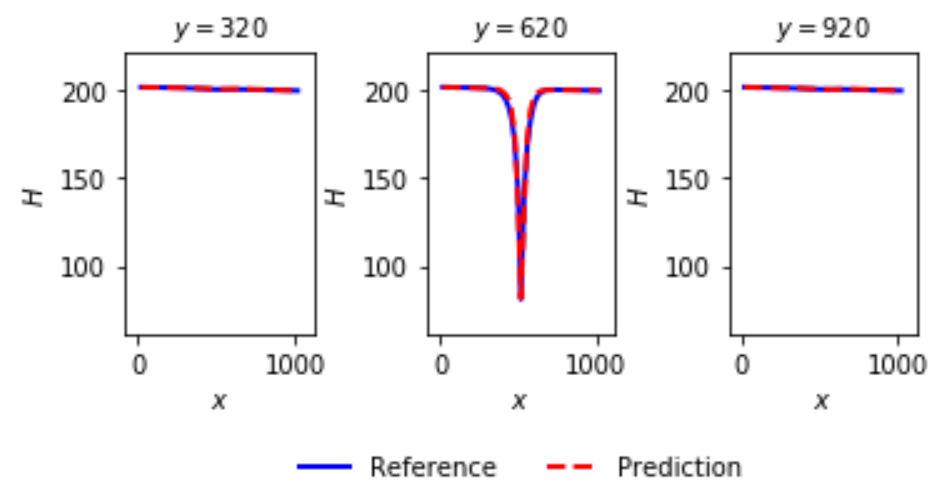

TgNN with EC

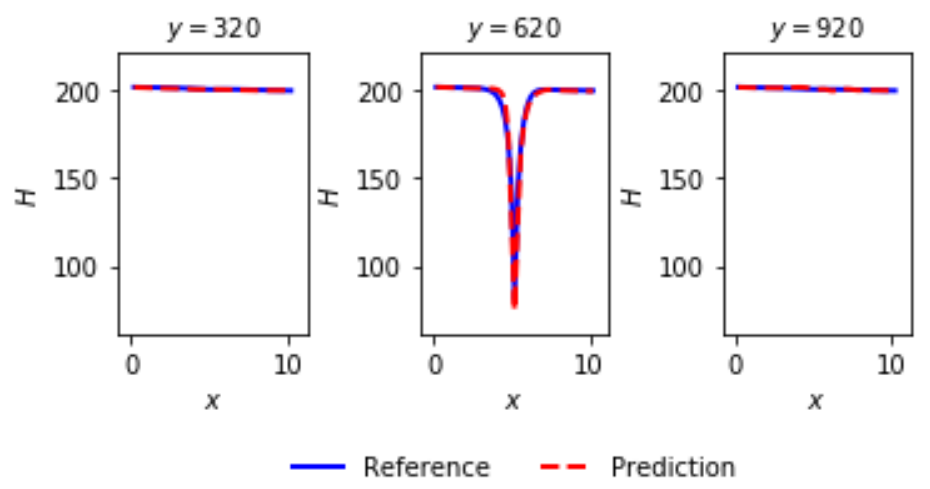

TgNN without EC

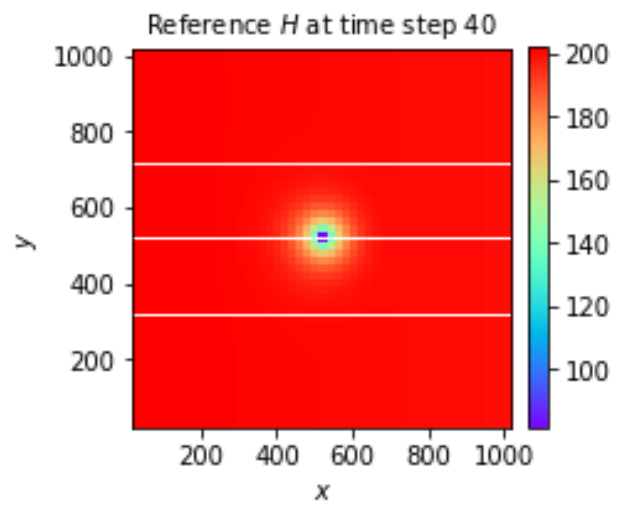

Reference

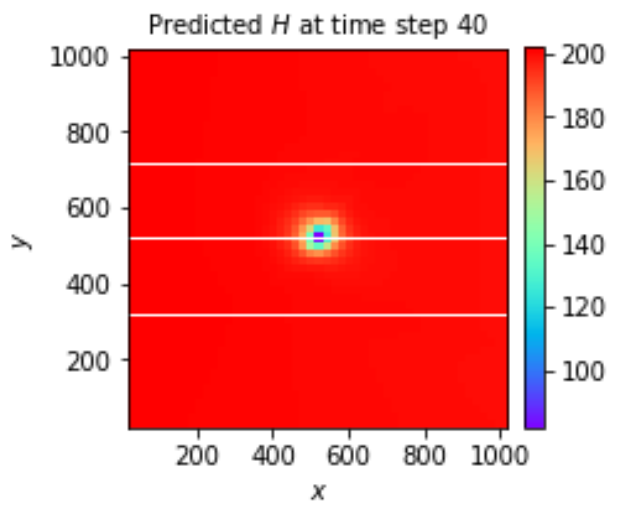

TgNN with EC

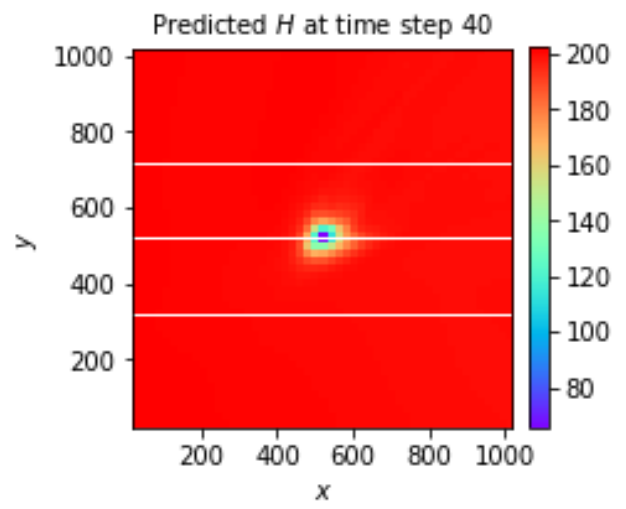

TgNN without EC

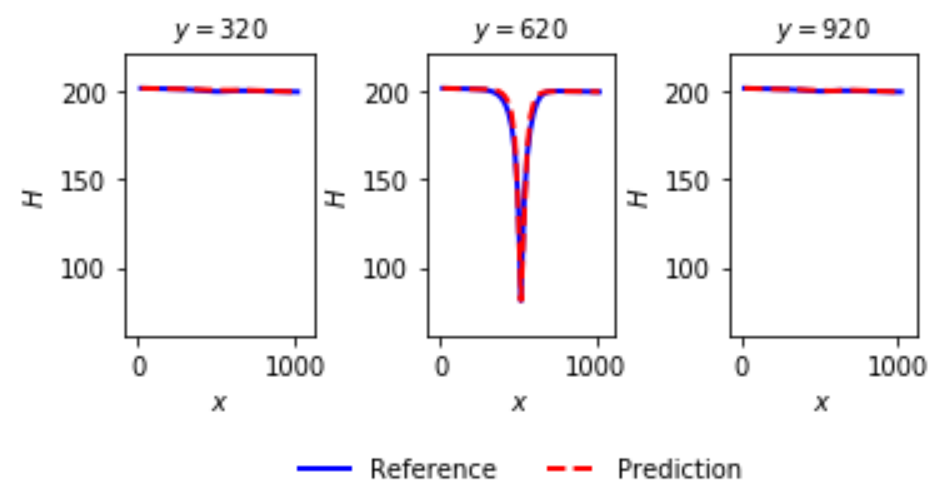

TgNN with EC

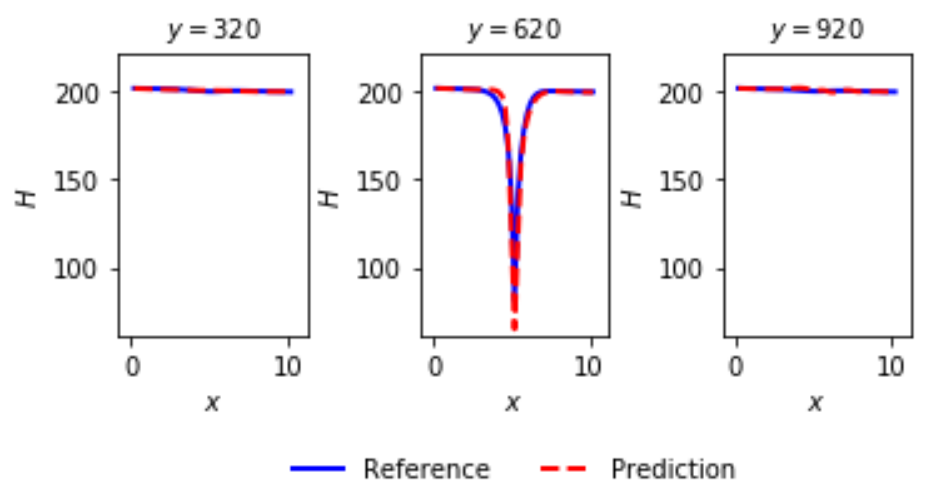

TgNN without EC

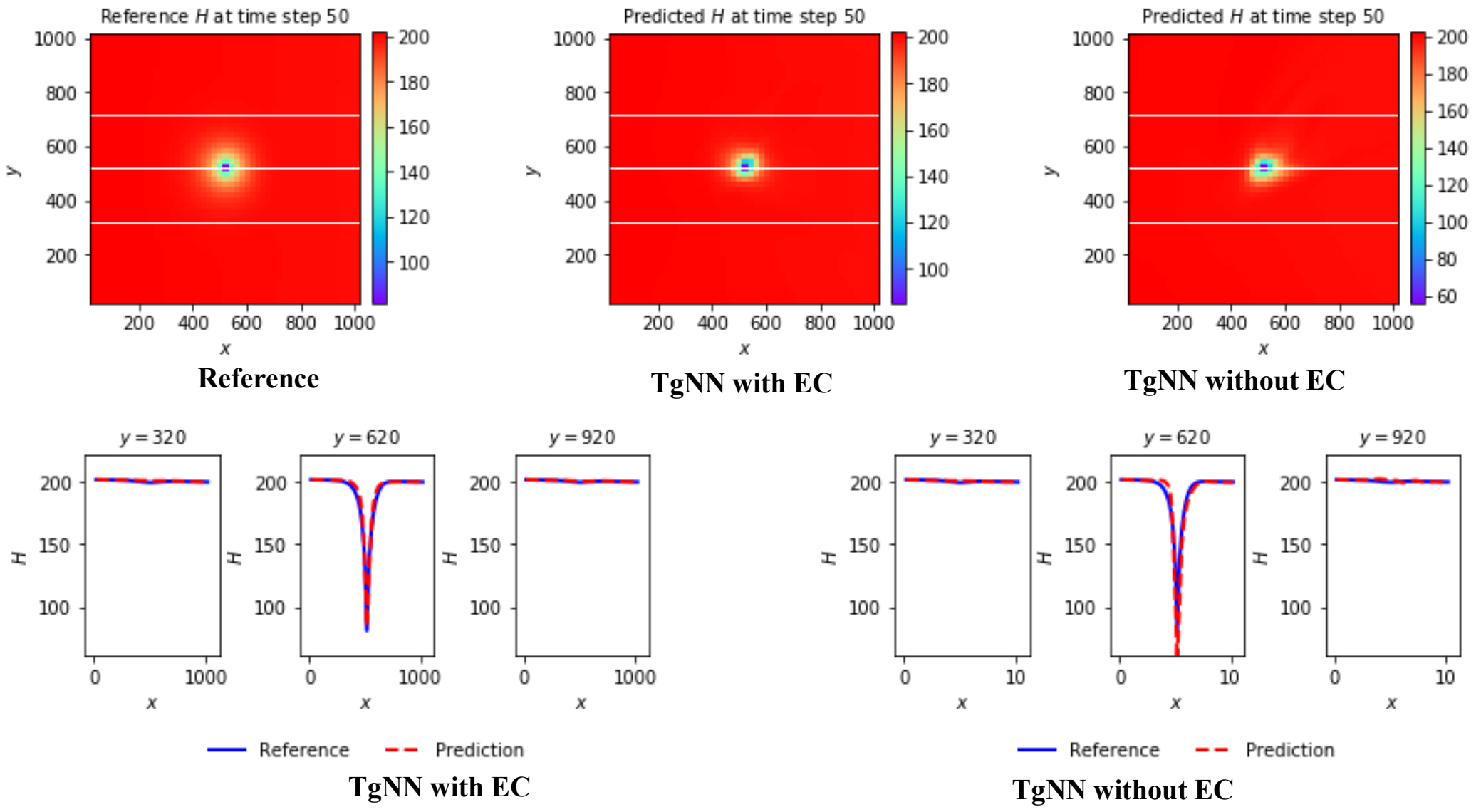

**Figure C.1.** Predictions of hydraulic head through TgNN with and without engineering control.